\documentclass[sigconf,authorversion,screen]{acmart}

\DeclareCaptionFont{bf}{\bfseries\boldmath}
\acmConference[ARES 2019]{14th International Conference on Availability, Reliability and Security}{August 26--29, 2019}{University of Kent, Canterbury, UK}

\usepackage[utf8]{inputenc}
\usepackage{amsmath}
\usepackage{bm,array}
\usepackage{booktabs}
\usepackage{caption}
\usepackage{float}
\usepackage{graphicx,dblfloatfix}
\usepackage{multirow}
\usepackage{url}
\usepackage{xspace}
\usepackage{enumitem}
\usepackage{titlesec}
\usepackage{calc}
\usepackage{enumitem}

\renewcommand*{\AA}     {AA\xspace}

\renewcommand*{\aa}     {authorship attribution\xspace}

\newcommand*{\AO}       {AO\xspace}

\newcommand*{\AV}       {AV\xspace}

\newcommand*{\av}       {authorship verification\xspace}

\newcommand*{\A}            {\mathcal{A}}
\newcommand*{\notA}         {\neg\A}
\newcommand*{\unknown}      {\mathcal{U}} 
\newcommand*{\Aunk}         {\unknown}

\newcommand*{\D}            {\mathcal{D}}
\newcommand*{\DA}           {{\D_{\A}}}
\newcommand*{\Dknown}       {\D_{\A\vphantom{\Aunk}}}
\newcommand*{\Dunk}         {\D_{\,\unknown}}

\newcommand*{\Dset}         {\mathbb{D}}

\newcommand*{\Corpus}       {\mathcal{C}}

\newcommand*{\Model}        {\mathcal{M}}

\newcommand*{\Arefset}      {\Dset_{\A}}

\newcommand*{\Problem}        {\rho}

\newcommand*{\Threshold}      {\theta}
\newcommand*{\ThresholdModel} {\theta_{\Model}}

\newcommand*{\CorpusDBLP}   {\Corpus_\mathrm{DBLP}}
\newcommand*{\CorpusReddit} {\Corpus_\mathrm{Reddit}}
\newcommand*{\CorpusPeeJ}   {\Corpus_\mathrm{Perv}}

\newcommand*{\classY}      {\textup{\ttfamily Y}\xspace} 
\newcommand*{\classN}      {\textup{\ttfamily N}\xspace}

\newcommand*{\accuracy}    {Accuracy\xspace}
\newcommand*{\fOne}        {F$_{1}$\xspace}

\newcommand*{\auc}         {AUC\xspace}

\newcommand*{\catOne}      {c@1\xspace}
\newcommand*{\auccat}      {AUC@1\xspace} 

\newcommand*{\bagnall}         {\textsf{Caravel}\xspace}
\newcommand*{\occav}           {\textsf{OCCAV}\xspace}
\newcommand*{\mocc}            {\textsf{MOCC}\xspace}
\newcommand*{\aveer}           {\textsf{AVeer}\xspace}
\newcommand*{\coav}            {\textsf{COAV}\xspace}
\newcommand*{\veenmanNNCD}     {\textsf{NNCD}\xspace}
\newcommand*{\glad}            {\textsf{GLAD}\xspace}
\newcommand*{\cng}             {\textsf{CNG}\xspace}
\newcommand*{\koppelGI}        {\textsf{GenIM}\xspace}

\newcommand*{\koppelUnmask}    {\textsf{Unmasking}\xspace}

\newcommand*{\noeckerDist}     {\textsf{DistAV}\xspace}
\newcommand*{\spatium}         {\textsf{SPATIUM}\xspace}

\newcommand*{\stamatatosImpGI} {\textsf{ImpGI}\xspace}

\newcommand{\e}[1]{\emph{#1}}

\newcommand*{\stateOfTheArt} {state of the art\xspace}

\newcommand*{\numberOfAVmethods} {12\xspace}

\newcommand*{\eg}            {e.\,g.,\mbox{}\xspace}
\newcommand*{\ie}            {i.\,e.,\mbox{}\xspace}

\begin{document}
\title[Applicable Authorship Verification]{Assessing the Applicability of Authorship Verification Methods}


\author{Oren Halvani}
\authornote{Corresponding author.}
\author{Christian Winter}
\author{Lukas Graner}
\affiliation{%
	\institution{Fraunhofer Institute for Secure Information Technology SIT, Darmstadt, Germany}
}
\email{<FirstName>.<LastName>@sit.fraunhofer.de}

\begin{abstract}
Authorship verification (AV) is a research subject in the field of digital text forensics that concerns itself with the question, whether two documents have been written by the same person. During the past two decades, an increasing number of proposed \AV approaches can be observed. However, a closer look at the respective studies reveals that the underlying characteristics of these methods are rarely addressed, which raises doubts regarding their applicability in real forensic settings. The objective of this paper is to fill this gap by proposing clear criteria and properties that aim to improve the characterization of existing and future AV approaches. Based on these properties, we conduct three experiments using \numberOfAVmethods existing AV approaches, including the current \stateOfTheArt. The examined methods were trained, optimized and evaluated on three self-compiled corpora, where each corpus focuses on a different aspect of applicability. Our results indicate that part of the methods are able to cope with very challenging verification cases such as 250 characters long informal chat conversations (72.7\% accuracy) or cases in which two scientific documents were written at different times with an average difference of 15.6 years (> 75\% accuracy). However, we also identified that all involved methods are prone to cross-topic verification cases.
\end{abstract}


\copyrightyear{2019}
\acmYear{2019}
\setcopyright{acmcopyright}
\acmConference[ARES '19]{Proceedings of the 14th International Conference on Availability, Reliability and Security (ARES 2019)}{August 26--29, 2019}{Canterbury, United Kingdom}
\acmBooktitle{Proceedings of the 14th International Conference on Availability, Reliability and Security (ARES 2019) (ARES '19), August 26--29, 2019, Canterbury, United Kingdom}
\acmPrice{15.00}
\acmDOI{10.1145/3339252.3340508}
\acmISBN{978-1-4503-7164-3/19/08}

\maketitle
\section{Introduction} \label{Introduction}
Digital text forensics aims at examining the originality and credibility of information in electronic documents and, in this regard, at extracting and analyzing information about the authors of the respective texts \cite{DigitalTextForensicsECIR:2019}. Among the most important tasks of this field are \textbf{\aa} (\AA) and \textbf{\av} (\AV), where the former deals with the problem of identifying the most likely author of a document $\Dunk$ with unknown authorship, given a set of texts of candidate authors. \AV, on the other hand, focuses on the question whether $\Dunk$ was in fact written by a known author $\A$, where only a set of reference texts $\Arefset$ of this author is given. Both disciplines are strongly related to each other, as any \AA problem can be broken down into a series of \AV problems \cite{StamatatosPothaImprovedIM:2017}. Breaking down an \AA problem into multiple \AV problems is especially important in such scenarios, where the presence of the true author of $\Dunk$ in the candidate set cannot be guaranteed. 

In the past two decades, researchers from different fields including linguistics, psychology, computer science and mathematics proposed numerous techniques and concepts that aim to solve the \AV task. Probably due to the interdisciplinary nature of this research field, \AV approaches were becoming more and more diverse, as can be seen in the respective literature. 
In 2013, for example, Veenman and Li \cite{VeenmanPAN13:2013} presented an \AV method based on compression, which has its roots in the field of information theory. 
In 2015, Bagnall \cite{BagnallRNN:2015} introduced the first deep learning approach that makes use of language modeling, an important key concept in statistical natural language processing. 
In 2017, Casta{\~{n}}eda and Calvo \cite{CastanedaAVviaLDA:2017} proposed an \AV method that applies a semantic space model through \e{Latent Dirichlet Allocation}, a generative statistical model used in information retrieval and computational linguistics. 

Despite the increasing number of \AV approaches, a closer look at the respective studies reveals that only minor attention is paid to their underlying characteristics such as reliability and robustness. These, however, must be taken into account before \AV methods can be applied in real forensic settings. The objective of this paper is to fill this gap and to propose important properties and criteria that are not only intended to characterize \AV methods, but also allow their assessment in a more systematic manner. By this, we hope to contribute to the further development of this young\footnote{According to the literature \cite{PANOverviewAV:2014}, Stamatatos et al. were the first researchers who discussed \AV in the context of natural language texts in 2000 \cite{StamatatosAutomaticTextCategorization:2000}. \AV, therefore, can be seen as a young field in contrast to \AA, which dates back to the 19th century \cite{HolmesEvolutionStylometryHumanities:1998}.} research field. Based on the proposed properties, we investigate the applicability of \numberOfAVmethods existing \AV approaches on three self-compiled corpora, where each corpus involves a specific challenge. 

The rest of this paper is structured as follows. 
Section~\ref{RelatedWork} discusses the related work that served as an inspiration for our analysis. 
Section~\ref{Analysis} comprises the proposed criteria and properties to characterize \AV methods. 
Section~\ref{Methodology} describes the methodology, consisting of the used corpora, examined \AV methods, selected performance measures and experiments. 
Finally, Section~\ref{Conclusions} concludes the work and outlines future work.

\section{Related Work} \label{RelatedWork}
Over the years, researchers in the field of authorship analysis identified a number of challenges and limitations regarding existing studies and approaches. Azarbonyad et al. \cite{AzarbonyadTimeAwareAA:2015}, for example, focused on the questions if the writing styles of authors of short texts change over time and how this affects \AA. To answer these questions, the authors proposed an \AA approach based on time-aware language models that incorporate the temporal changes of the writing style of authors. In one of our experiments, we focus on a similar question, namely, whether it is possible to recognize the writing style of authors, despite of large time spans between their documents. However, there are several differences between our experiment and the study of Azarbonyad et al. First, the authors consider an \AA task, where one anonymous document $\Dunk$ has to be attributed to one of $n$ possible candidate authors, while we focus on an \AV task, where $\Dunk$ is compared against one document $\DA$ of a known author. Second, the authors focus on texts with informal language (emails and tweets) in their study, while in our experiment we consider documents written in a formal language (scientific works). Third, Azarbonyad et al. analyzed texts with a time span of four years, while in our experiment the average time span is 15.6 years. Fourth, in contrast to the approach of the authors, none of the \numberOfAVmethods examined \AV approaches in our experiment considers a special handling of temporal stylistic changes. 

In recent years, the new research field \textbf{author obfuscation} (\AO) evolved, which concerns itself with the task to fool \AA or \AV methods in a way that the true author cannot be correctly recognized anymore. To achieve this, \AO approaches which, according to Gr{\"{o}}ndahl and Asokan \cite{GrondahlTextanalysisAdversarial:2019} can be divided into manual, computer-assisted and automatic types, perform a variety of 
modifications on the texts. These include simple synonym replacements, rule-based substitutions or word order permutations. In 2016, Potthast et al. \cite{PANOverviewAO:2016} presented the first large-scale evaluation of three \AO approaches that aim to attack 44 \AV methods, which were submitted to the PAN-\AV competitions during 2013-2015 \cite{PANOverviewAV:2013,PANOverviewAV:2014,PANOverviewAV:2015}. One of their findings was that even basic \AO approaches have a significant impact on many \AV methods. More precisely, the best-performing \AO approach was able to flip on average $\approx47$\% of an authorship verifier’s decisions towards choosing \classN (``different author''), while in fact \classY (``same author'') was correct \cite{PANOverviewAO:2016}. In contrast to Potthast et al., we do not focus on \AO to measure the robustness of \AV methods. Instead, we investigate in one experiment the question how trained \AV models behave, if the lengths of the questioned documents are getting shorter and shorter. To our best knowledge, this question has not been addressed in previous \av studies.

\section{Characteristics of Authorship Verification} \label{Analysis}
Before we can assess the applicability of \AV methods, it is important to understand their fundamental characteristics. Due to the increasing number of proposed \AV approaches in the last two decades, the need arose to develop a systematization including the conception, implementation and evaluation of \av methods. In regard to this, only a few attempts have been made so far. In 2004, for example, Koppel and Schler \cite{KoppelAVOneClassClassification:2004} described for the first time the connection between \AV and unary classification, also known as \textbf{one-class classification}. In 2008, Stein et al. \cite{SteinMetaAnalysisAV:2008} compiled an overview of important algorithmic building blocks for \AV where, among other things, they also formulated three \AV problems as decision problems. In 2009, Stamatatos \cite{StamatatosSurvey:2009} coined the phrases \e{profile-} and \e{instance-based} approaches that initially were used in the field of \AA, but later found their way also into \AV. In 2013 and 2014, Stamatatos et al. \cite{PANOverviewAV:2013,StamatatosProfileCNG:2014} introduced the terms \e{intrinsic-} and \e{extrinsic} models that aim to further distinguish between \AV methods. However, a closer look at previous attempts to characterize \av approaches reveals a number of misunderstandings, for instance, when it comes to draw the borders between their underlying classification models. In the following subsections, we clarify these misunderstandings, where we redefine previous definitions and propose new properties that enable a better comparison between \AV methods.

\subsection{Reliability (Determinism)} \label{AV_Property_Deterministic_Stochastic_Methods}
Reliability is a fundamental property any \AV method must fulfill in order to be applicable in real-world forensic settings. However, since there is no consistent concept nor a uniform definition of the term ``\textbf{reliability}'' in the context of \av according to the screened literature, we decided to reuse a definition from applied statistics, and adapt it carefully to \AV. 

In his standard reference\footnote{According to \e{Google Scholar}, Bollen's book was cited more than 30,000 times making it a standard reference across different research fields.} book, Bollen \cite{BollenReliability:1989} gives a clear description for this term: ``\e{Reliability is the consistency of measurement}'' and provides a simple example to illustrate its meaning: At time $t_1$ we ask a large number of persons the same question \e{Q} and record their responses. Afterwards, we remove their memory of the dialogue. At time $t_2$ we ask them again the same question \e{Q} and record their responses again. ``\e{The reliability is the consistency of the responses across individuals for the two time periods. To the extent that all individuals are consistent, the measure is reliable}'' \cite{BollenReliability:1989}. This example deals with the consistency of the measured objects as a factor for the reliability of measurements. In the case of \av, the analyzed objects are static data, and hence these cannot be a source of inconsistency. However, the measurement system itself can behave inconsistently and hence unreliable. This aspect can be described as \textbf{intra-rater reliability}.


Reliability in \av is satisfied, if an \AV method always generates the same prediction $\alpha \in \{ \classY, \classN \}$ for the same input $\Problem = (\Dunk, \DA)$, or in other words, if the method behaves \textbf{deterministically}. Several \AV approaches, including \cite{HalvaniARES:2014,HalvaniARES:2017,HalvaniDFRWS:2016,HalvaniOCCAV:2018,GLAD:2015,NoeckerDistractorlessAV:2012,JankowskaCNGAV:2013,JankowskaAVviaCNG:2014,StamatatosProfileCNG:2014} fall into this category. In contrast, if an \AV method behaves \textbf{non-deterministically} such that two different predictions for $\Problem$ are possible, the method can be rated as unreliable. Many \AV approaches, including  \cite{CastanedaAVviaLDA:2017,KoppelAVOneClassClassification:2004,NealAVviaIsolationForests:2018,StamatatosPothaImprovedIM:2017,SeidmanPAN13:2013,BagnallRNN:2015,BevendorffUnmasking:2019,HernandezHomotopyAV:2015,KocherPANSpatium:2015} belong to this category, since they involve randomness (\eg weight initialization, feature subsampling, chunk generation or impostor selection), which might distort the evaluation, as every run on a test corpus very likely leads to different results. Under lab conditions, results of non-deterministic \AV methods can (and should) be counteracted by averaging multiple runs. However, it remains highly questionable if such methods are generally applicable in realistic forensic cases, where the prediction $\alpha$ regarding a verification case $\Problem$ might sometimes result in \classY and sometimes in \classN.

\subsection{Optimizability}
Another important property of an \AV method is optimizability. We define an \AV method as \textbf{optimizable}, if it is designed in such a way that it offers adjustable hyperparameters that can be tuned against a training/validation corpus, given an optimization method such as grid or random search. Hyperparameters might be, for instance, the selected distance/similarity function, the number of layers and neurons in a neural network or the choice of a kernel method. The majority of existing \AV approaches in the literature (for example, \cite{KoppelAVOneClassClassification:2004,NoeckerDistractorlessAV:2012,JankowskaCNGAV:2013,GLAD:2015,CastroAVAverageSimilarity:2015,CastanedaAVviaLDA:2017,KoppelWinter2DocsBy1:2014,StamatatosProfileCNG:2014}) belong to this category. On the other hand, if a published \AV approach involves hyperparameters that have been entirely fixed such that there is no further possibility to improve its performance from outside (without deviating from the definitions in the publication of the method), the method is considered to be \textbf{non-optimizable}. Non-optimizable \AV methods are preferable in forensic settings as, here, the existence of a training/validation corpus is not always self-evident. Among the proposed \AV approaches in the respective literature, we identified only a small fraction \cite{HalvaniOCCAV:2018,VeenmanPAN13:2013,KocherPANSpatium:2015} that fall into this category.


\subsection{Model Category}
From a machine learning point of view, \av represents a unary classification problem \cite{GLAD:2015,KoppelAVOneClassClassification:2004,StamatatosProfileCNG:2014,PothaStamatatosTopicAV:2018,SteinMetaAnalysisAV:2008}. Yet, in the literature, it can be observed that sometimes \AV is treated as a unary \cite{JankowskaAVviaCNG:2014,NoeckerDistractorlessAV:2012,NealAVviaIsolationForests:2018,StamatatosProfileCNG:2014} and sometimes as a binary classification task \cite{KocherPANSpatium:2015,KoppelWinter2DocsBy1:2014,GLAD:2015,VeenmanPAN13:2013}. We define the way an \AV approach is modeled by the phrase \textbf{model category}. However, before explaining this in more detail, we wish to recall what unary/one-class classification exactly represents. For this, we list the following verbatim quotes, which characterize one-class classification, as can be seen, almost identically (emphasis by us):
\begin{itemize}
	\item \e{``In one-class classification it is assumed that only information of one of the classes, the target class, is available. This means that \textbf{just example objects of the target class} can be used and that \textbf{no information} about the other class \textbf{of outlier objects is present}.''} \cite{TaxOCC:2001}

	\item \e{``One-class classification (OCC) [\dots] consists in making a description of a target class of objects and in detecting whether a new object resembles this class or not. [\dots] The OCC model is developed \textbf{using target class samples only}.''} \cite{RodionovaOCC:2016}

	\item \e{``In one-class classification framework, an object is classified as belonging or not belonging to a target class, while \textbf{only} sample \textbf{examples} of objects \textbf{from the target class} are available during the training phase.''} \cite{JankowskaAVviaCNG:2014}
\end{itemize}
Note that in the context of \av, target class refers to the known author $\A$ such that for a document $\Dunk$ of an unknown author $\unknown$ the task is to verify whether $\unknown = \A$ holds. One of the most important requirements of any existing \AV method is a \textbf{decision criterion}, which aims to accept or reject a questioned authorship. A decision criterion can be expressed through a simple scalar threshold $\Threshold$ or a more complex model $\ThresholdModel$ such as a hyperplane in a high-dimensional feature space. As a consequence of the above statements, the determination of $\Threshold$ or $\ThresholdModel$ has to be performed solely on the basis of $\Arefset$, otherwise the \AV method cannot be considered to be unary. However, our conducted literature research regarding existing \AV approaches revealed that there are uncertainties how to precisely draw the borders between unary and binary \AV methods (for instance, \cite{BoukhaledProbabilisticAV:2014,StamatatosProfileCNG:2014,PothaStamatatosTopicAV:2018}). Nonetheless, few attempts have been made to distinguish both categories from another perspective.
Potha and Stamatatos \cite{PothaStamatatosTopicAV:2018}, for example, categorize \AV methods as either \textbf{intrinsic} or \textbf{extrinsic} (emphasis by us):
\begin{enumerate}
	\item \e{``Intrinsic verification models view it [\ie the verification task] as a one-class classification task and are based exclusively on analysing the similarity between [$\Arefset$] and [$\Dunk$]. [\dots] Such methods [\dots] \textbf{do not require any external resources}.''} \cite{PothaStamatatosTopicAV:2018}

	\item \e{``On the other hand, extrinsic verification models attempt to transform the verification task to a pair classification task by \textbf{considering external documents} to be used as \textbf{samples of the negative class}.}'' \cite{PothaStamatatosTopicAV:2018}
\end{enumerate}
While we agree with statement (2), the former statement (1) is unsatisfactory, as intrinsic verification models are \textbf{not necessarily unary}. For example, the \AV approach GLAD proposed by H{\"u}rlimann et al. \cite{GLAD:2015} directly contradicts statement (1). Here, the authors \e{``decided to cast the problem as a \textbf{binary classification task} where class values are \classY [$\A = \unknown$] and \classN [$\A \neq \unknown$]. [\dots] We do \textbf{not introduce any negative examples} by means of external documents, thus adhering to an \textbf{intrinsic approach}.''} \cite{GLAD:2015}.

A misconception similar to statement (1) can be observed in the paper of Jankowska et al. \cite{JankowskaCNGAV:2013}, who introduced the so-called \cng approach claimed to be a one-class classification method. \cng is intrinsic in that way that it considers only $\Arefset$ when deciding a problem $\Problem$. However, the decision criterion, which is a threshold $\Threshold$, is determined on a set of verification problems, labeled either as \classY or \classN. This incorporates ``external resources'' for defining the decision criterion, and it constitutes an implementation of binary classification between \classY and \classN in analogy to the statement of Hürlimann et al. \cite{GLAD:2015} mentioned above. Thus, \cng is in conflict with the unary definition mentioned above. In a subsequent paper \cite{JankowskaAVviaCNG:2014}, however, Jankowska et al. refined their approach and introduced a modification, where $\Threshold$ was determined solely on the basis of $\Arefset$. Thus, the modified approach can be considered as a true unary \AV method, according to the quoted definitions for unary classification.

In 2004, Koppel and Schler \cite{KoppelAVOneClassClassification:2004} presented the \koppelUnmask approach which, according to the authors, represents a unary \AV method. However, if we take a closer look at the learning process of \koppelUnmask, we can see\footnote{See the intuitive illustration provided in \cite[Figure 1]{BevendorffUnmasking:2019}.} that it is based on a binary SVM classifier that consumes feature vectors (derived from \e{``degradation curves''}) labeled as \classY (``same author'') or \classN (``different author''). \koppelUnmask, therefore, cannot be considered to be unary as the decision is not solely based on the documents within $\Arefset$, in analogy to the \cng approach of Jankowska et al. \cite{JankowskaCNGAV:2013} discussed above. 

\begin{figure*}
	\centering
	\includegraphics[scale=0.54]{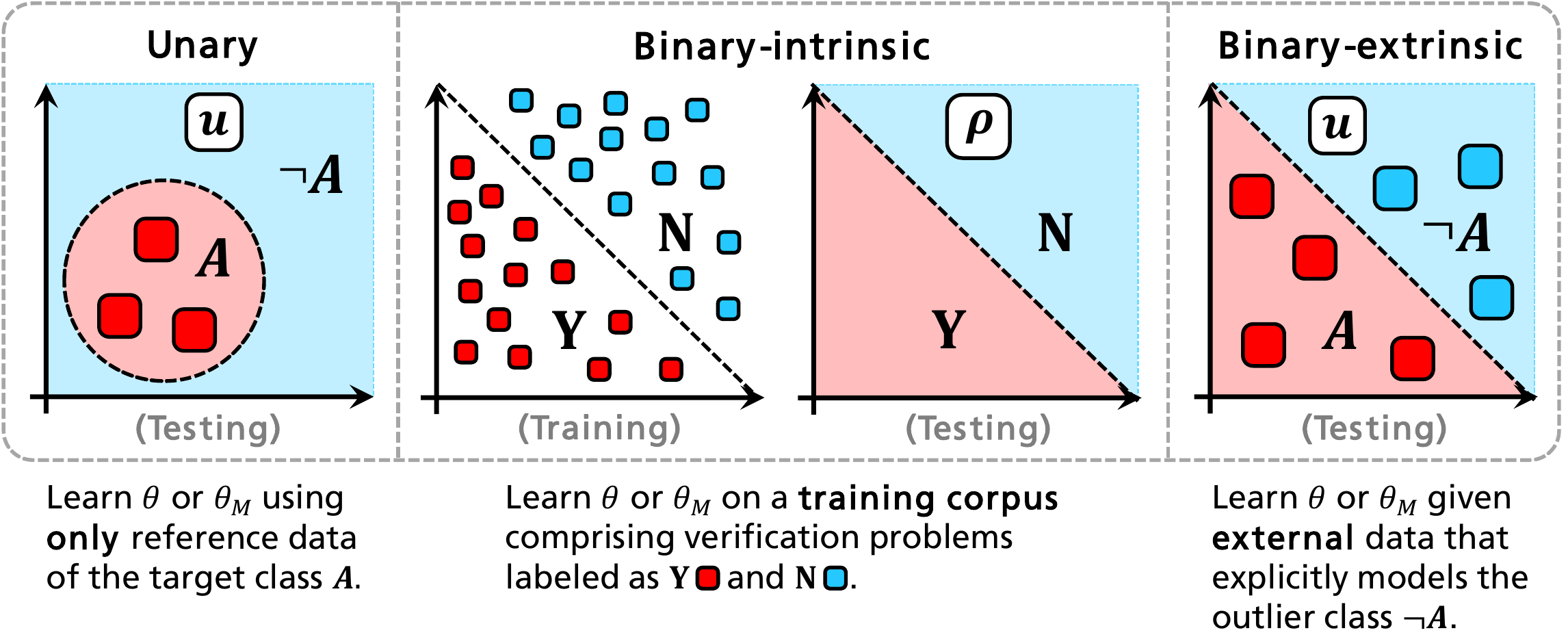}
	\caption{The three possible model categories of \av approaches. Here, $\unknown$ refers to the instance (for example, a document or a feature vector) of the unknown author. $\A$ is the target class (known author) and $\notA$ the outlier class (any other possible author). In the binary-intrinsic case, $\Problem$ denotes the verification problem (subject of classification), and \classY and \classN denote the regions of the problem feature space where, according to a training corpus, the authorship holds or not. \label{fig:AV_Model_Categories}}
\end{figure*}

It should be highlighted again that the aforementioned three approaches are \textbf{binary-intrinsic} since their decision criteria $\Threshold$ or $\ThresholdModel$ was determined on a set of problems labeled in a binary manner (\classY and \classN) while after training, the verification is performed in an intrinsic manner, meaning that $\Arefset$ and $\Dunk$ are compared against $\Threshold$ or $\ThresholdModel$ but not against documents within other verification problems (cf. Figure~\ref{fig:AV_Model_Categories}). A crucial aspect, which might have lead to misperceptions regarding the model category of these approaches in the past, is the fact that two different class domains are involved. On the one hand, there is the \textbf{class domain of authors}, where the task is to distinguish $\A$ and $\notA$. On the other hand, there is the \emph{elevated} or \emph{lifted} \textbf{domain of verification problem classes}, which are \classY and \classN. The training phase of binary-intrinsic approaches is used for learning to distinguish these two classes, and the verification task can be understood as putting the verification problem as a whole into class \classY or class \classN, whereby the class domain of authors fades from the spotlight (cf.\ Figure~\ref{fig:AV_Model_Categories}).

Besides unary and binary-intrinsic methods, there is a third category of approaches, namely \textbf{binary-extrinsic} \AV approaches (for example,  \cite{BagnallRNN:2015,KocherPANSpatium:2015,HernandezHomotopyAV:2015,KhonjiIraqiAV:2014,KoppelWinter2DocsBy1:2014,StamatatosPothaImprovedIM:2017,VeenmanPAN13:2013}). These methods use external documents during a potentially existing training phase and -- more importantly -- during testing. In these approaches, the decision between $\A$ and $\notA$ is put into the focus, where the external documents aim to construct the counter class $\notA$. 

Based on the above observations, we conclude that the key requirement for judging the model category of an \AV method depends solely on the aspect \textbf{how} its decision criterion $\Threshold$ or $\ThresholdModel$ is determined (cf.\ Figure~\ref{fig:AV_Model_Categories}):
\begin{enumerate}
	\item An \AV method is \textbf{unary} if and only if its decision criterion $\Threshold$ or $\ThresholdModel$ is determined solely on the basis of the target class $\A$ during testing. As a consequence, an \AV method cannot be considered to be unary if documents not belonging to $\A$ are used to define $\Threshold$ or $\ThresholdModel$.

	\item An \AV method is \textbf{binary-intrinsic} if its decision criterion $\Threshold$ or $\ThresholdModel$ is determined on a training corpus comprising verification problems labeled either as \classY or \classN (in other words documents of several authors). However, once the training is completed, a binary-intrinsic method has no access to external documents anymore such that the decision regarding the authorship of $\Dunk$ is made on the basis of the reference data of $\A$ as well as $\Threshold$ or $\ThresholdModel$.

	\item An \AV method is \textbf{binary-extrinsic} if its decision criterion $\Threshold$ or $\ThresholdModel$ is determined during testing on the basis of external documents that represent the outlier class $\notA$. 
\end{enumerate}

Note that optimizable \AV methods such as \cite{HalvaniARES:2014,JankowskaAVviaCNG:2014} are not excluded to be unary. Provided that $\Threshold$ or $\ThresholdModel$ is not subject of the optimization procedure, the model category remains unary. The reason for this is obvious; \textbf{Hyperparameters} might influence the resulting performance of unary \AV methods. The \textbf{decision criterion} itself, however, remains unchanged.

\subsection{Implications}
Each model category has its own implications regarding prerequisites, evaluability, and applicability.

\subsubsection{Unary AV Methods:} \label{Implications_UnaryAV}
One advantage of unary \AV methods is that they do not require a specific document collection strategy to construct the counter class $\neg\A$, which reduces their complexity. On the downside, the choice of the underlying machine learning model of a unary \AV approach is restricted to one-class classification algorithms or unsupervised learning techniques, given a suitable decision criterion. 

However, a far more important implication of unary \AV approaches concerns their \textbf{performance assessment}. Since unary classification (not necessarily \AV) approaches depend on a fixed decision criterion $\Threshold$ or $\ThresholdModel$, performance measures such as the area under the ROC curve (\auc) are meaningless. Recall that ROC analysis is used for evaluating classifiers, where the decision threshold is not finally fixed. ROC analysis requires that the classifier generates scores, which are comparable across classification problem instances. The ROC curve and the area under this curve is then computed by considering all possible discrimination thresholds for these scores. While unary \AV approaches might produce such scores, introducing a \textbf{variable} $\Threshold$ would change the semantics of these approaches. Since unary \AV approaches have a fixed decision criterion, they provide only a single point in the ROC space. To assess the performance of a unary \AV method, it is, therefore, mandatory to consider the confusion matrix that leads to this point in the ROC space. 

Another implication is that unary \AV methods are necessarily instance-based and, thus, require a set $\Arefset = \{ \D_1, \D_2, \ldots \}$ of multiple  documents of the known author $\A$. If only one reference document is available ($\Arefset = \{\DA\} $), this document must be artificially turned into multiple samples from the author. In general, unary classification methods need multiple samples from the target class since it is not possible to determine a \emph{relative} closeness to that class based on only one sample.

\subsubsection{Binary AV Methods:}
On the plus side, binary-intrinsic or extrinsic \AV methods benefit from the fact that we can choose among a variety of \textbf{binary}\footnote{For example: Support vector machines, logistic regression or perceptrons.} and $\bm{n}$\textbf{-ary}\footnote{For example: Naive Bayes, random forests or a variety of neural networks.} classification models. However, if we consider designing a binary-intrinsic \AV method, it should not be overlooked that the involved classifier will learn nothing about individual authors, but only similarities or differences that hold in general for \classY and \classN verification problems \cite{KoppelWinter2DocsBy1:2014}.

If, on the other hand, the choice falls on a binary-extrinsic method, a strategy has to be considered for collecting representative documents for the outlier class $\notA$. Several existing methods such as \cite{KoppelWinter2DocsBy1:2014,StamatatosPothaImprovedIM:2017,VeenmanPAN13:2013} rely on search engines for retrieving appropriate documents, but these search engines might refuse their service if a specified quota is exhausted. Additionally, the retrieved documents render these methods inherently \textbf{non-deterministic}. Moreover, such methods cause relatively \textbf{high runtimes} \cite{PANOverviewAV:2013,PANOverviewAV:2014}. Using search engines also requires an active Internet connection, which might not be available or allowed in specific scenarios. But even if we can access the Internet to retrieve documents, there is \textbf{no guarantee} that the true author is not among them. With these points in mind, the \textbf{applicability} of binary-extrinsic methods in real-world cases, \ie in real forensic settings, remains highly questionable.

\section{Methodology} \label{Methodology}
In the following, we introduce our three self-compiled corpora, where each corpus represents a different challenge. Next, we describe which \av approaches we considered for the experiments and classify each \AV method according to the properties introduced in Section~\ref{Analysis}. Afterwards, we explain which performance measures were selected with respect to the conclusion made in Section~\ref{Implications_UnaryAV}. Finally, we describe our experiments, present the results and highlight a number of observations.

\subsection{Corpora} \label{Corpora}
A serious challenge in the field of \AV is the lack of publicly available (and suitable) corpora, which are required to train and evaluate \AV methods. Among the few publicly available corpora are those that were released by the organizers of the well-known PAN-\AV competitions\footnote{\url{https://pan.webis.de}} \cite{PANOverviewAV:2013,PANOverviewAV:2014,PANOverviewAV:2015}. In regard to our experiments, however, we cannot use these corpora, due to the absence of relevant meta-data such as the precise time spans where the documents have been written as well as the topic category of the texts. Therefore, we decided to compile our own corpora based on English documents, which we crawled from different publicly accessible sources. In what follows, we describe our three constructed corpora, which are listed together with their statistics in Table~\ref{tab:CorpusStatistics}. Note that all corpora are balanced such that verification cases with \textbf{matching} (\classY) and \textbf{non-matching} (\classN) authorships are evenly distributed. 
\begin{table}
	\begin{tabular}{lrrrr}
		\toprule
		\bfseries\boldmath Corpus $\Corpus$         & \boldmath$|\Corpus|$  & \boldmath$|\Arefset|$ & \bfseries\boldmath avg\_len($\Dunk$) & \bfseries\boldmath avg\_len($\DA$) \\\midrule
		$\CorpusDBLP$ (train)    & 32  & 1 & 6,771  & 5,491 \\
		$\CorpusDBLP$ (test)     & 48  & 1 & 7,650  & 5,714 \\\midrule
		$\CorpusPeeJ$ (train)    & 440 & 2 & 8,157  & 7,268 \\ 
		$\CorpusPeeJ$ (test)     & 660 & 2 & 9,611 & 8,692 \\\midrule
		$\CorpusReddit$ (train)  & 40  & 1 & 7,011  & 6,909 \\
		$\CorpusReddit$ (test)   & 60  & 1 & 6,974  & 6,990 \\		
		\bottomrule
	\end{tabular}
	\caption{All training and testing corpora used in our experiments. Here, $|\Corpus|$ denotes the number of verification problems in each corpus $\Corpus$ and $|\Arefset|$ the number of the known documents. The average character length of the unknown document $\Dunk$ and the known document $\DA$ (concatenation of all known documents $\Arefset$) is denoted by avg\_len($\Dunk$) and avg\_len($\DA$), respectively. \label{tab:CorpusStatistics}}
\end{table}

\subsubsection{DBLP Corpus} \label{Corpora_DBLP}
As a first corpus, we compiled $\CorpusDBLP$ that represents a collection of 80 excerpts from scientific works including papers, dissertations, book chapters and technical reports, which we have chosen from the well-known \e{Digital Bibliography \& Library Project} (DBLP) platform\footnote{\url{https://dblp.uni-trier.de}}. Overall, the documents\footnote{Note that each document is single-authored.} were written by 40 researchers, where for each author $\A$, there are exactly two documents. Given the 80 documents, we constructed for each author $\A$ two verification problems $\Problem_1$ (a \classY-case) and $\Problem_2$ (an \classN-case). For $\Problem_1$ we set $\A$'s first document as $\DA$ and the second document as $\Dunk$. For $\Problem_2$ we reuse $\DA$ from $\Problem_1$ as the known document and selected a text from another (random) author as the unknown document. The result of this procedure is a set of 80 verification problems, which we split into a training and test set based on a 40/60\% ratio. Where possible, we tried to \textbf{restrict} the content of each text to the \textbf{abstract} and \textbf{conclusion} of the original work. However, since in many cases these sections were too short, we also considered other parts of the original works such as introduction or discussion sections. To ensure that the extracted text portions are appropriate for the \AV task, each original work was \textbf{preprocessed manually}. More precisely, we removed tables, formulas, citations, quotes and sentences that include non-language content such as mathematical constructs or specific names of researchers, systems or algorithms. The average time span between both documents of an author is 15.6 years. The minimum and maximum time span are 6 and 40 years, respectively. Besides the temporal aspect of $\CorpusDBLP$, another challenge of this corpus is the formal (scientific) language, where the usage of stylistic devices\footnote{For example, repetitions, metaphors, rhetorical questions, oxymorons, etc.} is more restricted, in contrast to other genres such as novels or poems.

\subsubsection{Perverted Justice Corpus} \label{Corpora_PeeJ}
As a second corpus, we compiled $\CorpusPeeJ$, which represents a collection of 1,645 chat conversations of 550 sex offenders crawled from the \e{Perverted-Justice} portal\footnote{\url{http://www.perverted-justice.com}}. The chat conversations stem from a variety of sources including emails and instant messengers (\eg \e{MSN}, \e{AOL} or \e{Yahoo}), where for each conversation, we ensured that only chat lines from the offender were extracted. We applied the same problem construction procedure as for the corpus $\CorpusDBLP$, which resulted in 1,100 verification problems that again were split into a training and test set given a 40/60\% ratio. In contrast to the corpus $\CorpusDBLP$, we only performed slight preprocessing. Essentially, we removed user names, time-stamps, URLs, multiple blanks as well as annotations that were not part of the original conversations from all chat lines. Moreover, we did not normalize words (for example, shorten words such as \e{``nooooo''} to \e{``no''}) as we believe that these represent important style markers. Furthermore, we did not remove newlines between the chat lines, as the positions of specific words might play an important role regarding the individual's writing style.

\subsubsection{Reddit Corpus} \label{Corpus_Reddit}
As a third corpus, we compiled $\CorpusReddit$, which is a collection of 200 aggregated postings crawled from the \e{Reddit} platform\footnote{\url{https://www.reddit.com}}. Overall, the postings were written by 100 Reddit users and stem from a variety of subreddits. In order to construct the \classY-cases, we selected exactly two postings from disjoint subreddits for each user such that both the known and unknown document $\DA$ and $\Dunk$ differ in their topic. Regarding the \classN-cases, we applied the opposite strategy such that $\DA$ and $\Dunk$ belong to the same topic. The rationale behind this is to figure out to which extent \AV methods can be fooled in cases, where the topic matches but not the authorship and vice versa. Since for this specific corpus we have to control the topics of the documents, we did not perform the same procedure applied for $\CorpusDBLP$ and $\CorpusPeeJ$ to construct the training and test sets. Instead, we used for the resulting 100 verification problems a 40/60\% hold-out split, where both training and test set are entirely disjoint.

\subsection{Examined Authorship Verification Methods} \label{AVMethods}
As a basis for our experiments, we reimplemented \numberOfAVmethods existing \AV approaches, which have shown their potentials in the previous PAN-\AV competitions \cite{PANOverviewAV:2013,PANOverviewAV:2015} as well as in a number of \AV studies. The methods are listed in Table~\ref{tab:AVMethods} together with their classifications regarding the \AV characteristics, which we proposed in Section~\ref{Analysis}. 
\begin{table}
	\small
	\begin{tabular}{llll} 	\toprule
		\textbf{AV Method}  & \textbf{Model Categ.}  & \textbf{Optimizability} & \textbf{Determinism}  \\ \midrule
		\mocc             \cite{HalvaniARES:2014} 					  & unary            & determin.      & optimizable      \\			
		\occav            \cite{HalvaniOCCAV:2018} 				      & unary            & determin.      & non-optimiz.\\ \midrule		
		\coav             \cite{HalvaniARES:2017}					  & binary-intr. & determin.     & optimizable     \\  
		\aveer            \cite{HalvaniDFRWS:2016}                    & binary-intr. & determin.     & optimizable      \\ 
		\glad             \cite{GLAD:2015}  						  & binary-intr. & determin.     & optimizable      \\  
		\noeckerDist      \cite{NoeckerDistractorlessAV:2012}  	      & binary-intr. & determin.     & optimizable      \\	
		\koppelUnmask     \cite{KoppelAVOneClassClassification:2004}  & binary-intr. & non-determin. & optimizable      \\ \midrule
		\bagnall          \cite{BagnallRNN:2015}                      & binary-extr. & non-determin.  & optimizable      \\
		\koppelGI         \cite{SeidmanPAN13:2013}                    & binary-extr. & non-determin.  & optimizable      \\
		\stamatatosImpGI  \cite{StamatatosPothaImprovedIM:2017}       & binary-extr. & non-determin.  & optimizable      \\
		\spatium          \cite{KocherPANSpatium:2015}                & binary-extr. & non-determin.  & non-optimiz. \\	
		\veenmanNNCD      \cite{VeenmanPAN13:2013}                    & binary-extr. & determin.	  & non-optimiz.  \\ \bottomrule		
	\end{tabular}
	\caption{All \numberOfAVmethods \AV methods, classified according to their properties. \label{tab:AVMethods}}
\end{table} 
All (optimizable) \AV methods were tuned regarding their hyperparameters, according to the original procedure mentioned in the respective paper. However, in the case of the binary-extrinsic methods (\koppelGI, \stamatatosImpGI and \veenmanNNCD) we had to use an alternative impostors generation strategy in our reimplementations, due to technical problems. In the respective papers, the authors used search engine queries to generate the impostor documents, which are needed to model the counter class $\neg\A$. Regarding our reimplementations, we used the documents from the static corpora (similarly to the idea of Kocher and Savoy \cite{KocherPANSpatium:2015}) to generate the impostors in the following manner: Let $\Corpus = \{ \rho_1, \rho_2, \ldots, \rho_n \}$ denote a corpus with $n$ verification problems. For each $\rho_i = ({\Dunk}_i, {\Arefset}_i)$ we choose all unknown documents ${\Dunk}_j$ in $\Corpus$ with $i \neq j$ and append them the impostor set $\mathbb{U}$. Here, it should be highlighted that both \koppelGI and \stamatatosImpGI consider the number of impostors as a \textbf{hyperparameter} such that the resulting impostor set is a subset of $\mathbb{U}$. In contrast to this, \veenmanNNCD considers all $\unknown_j \in \mathbb{U}$ as possible impostors. This fact plays an important role in the later experiments, where we compare the \AV approaches to each other. Although our strategy is not flexible like using a search engine, it has one advantage that, here, it is assumed that the true author of an unknown document is not among the impostors, since in our corpora the user/author names are known\footnote{However, it might be possible that behind multiple user names there is only one person (in other words, we cannot guarantee: one user = one account).} beforehand.

\subsection{Performance Measures} \label{PerformanceMeasures}
According to our extensive literature research, numerous measures (\eg \accuracy, \fOne, \catOne, \auc, \auccat, $\kappa$ or EER) have been used so far to assess the performance of \AV methods. In regard to our experiments, we decided to use \catOne and \auc for several reasons. First, \accuracy, \fOne and $\kappa$ are not applicable in cases where \AV methods leave verification problems unanswered, which concerns some of our examined \AV approaches. Second, using \auc alone is meaningless for non-optimizable \AV methods, as explained in Section~\ref{Implications_UnaryAV}. Third, both have been used in the PAN-\AV competitions \cite{PANOverviewAV:2014,PANOverviewAV:2015}. Note that we also list the confusion matrix outcomes.

\subsection{Experiments}
Overall, we focus on three experiments, which are based on the corpora introduced in Section~\ref{Corpora}: 
\begin{enumerate}
	\item The Effect of Stylistic Variation Across Large Time Spans	
	\item The Effect of Topical Influence	
	\item The Effect of Limited Text Length
\end{enumerate}
In the following each experiment is described in detail.

\subsubsection{The Effect of Stylistic Variation Across Large Time Spans:}
In this experiment, we seek to answer the question if the writing style of an author $\A$ can be recognized, given a large time span between two documents of $\A$. The motivation behind this experiment is based on the statement of Olsson \cite{ForensicLinguisticsBookOlsson:2008} that language acquisition is a \textbf{continuous process}, which is not only acquired, but also \textbf{can be lost}. Therefore, an important question that arises here is, if the writing style of a person remains ``stable'' across a large time span, given the fact that language in each individual's life is never ``fixed'' \cite{ForensicLinguisticsBookOlsson:2008}. Regarding this experiment, we used the $\CorpusDBLP$ corpus.  
\begin{table}	
	\begin{center}
		\begin{tabular}{lrrrrrrr} \toprule
			\textbf{AV Method} & \textbf{\catOne} & \textbf{\auc}  & \textbf{TP} & \textbf{FN} & \textbf{FP} & \textbf{TN}  &  \textbf{UP}  \\ \midrule			
			\bagnall          & \textbf{0.792}  & 0.905 &  19 & 5  & 5  & 19 & 0 \\
			\coav             & \textbf{0.750}  & 0.802 &  18 & 6  & 6  & 18 & 0 \\
			\veenmanNNCD      & 0.729  & {\color{gray} 0.858}  & 14 & 10 & 3  & 21 & 0 \\
			\spatium          & 0.717  & 0.788 &  15 & 5  & 5  & 14 & 9 \\    
			\glad             & 0.708  & \textbf{0.821} &  14 & 10 & 4  & 20 & 0 \\
			\stamatatosImpGI  & 0.708  & 0.816 &  22 & 2  & 12 & 12 & 0 \\
			\koppelGI         & 0.703  & 0.768 &  14 & 7  & 5  & 16 & 6 \\
			\koppelUnmask     & 0.688  & 0.747 &  19 & 5  & 10 & 14 & 0 \\
			\aveer            & 0.667  & 0.781 &  17 & 7  & 9  & 15 & 0 \\
			\noeckerDist      & 0.583  & {\color{gray} 0.681}  &  7 & 17  & 3  & 21 & 0 \\
			\mocc             & 0.542  & {\color{gray} 0.655}  &  7 & 17  & 5  & 19 & 0 \\
			\occav            & 0.521  & {\color{gray} 0.750}  &  1 & 23  & 0  & 24 & 0 \\ \bottomrule	
		\end{tabular}
	\end{center}
	\caption{Evaluation results for the test corpus $\CorpusDBLP$ in terms of \catOne and \auc. TP, FN, FP and TN represent the four confusion matrix outcomes, while UP denotes the number of unanswered verification problems. Note that AUC scores for the non-optimizable and unary AV methods are grayed out. \label{tab:DBLPEvaluationResults}}
\end{table} 
The results of the \numberOfAVmethods examined \AV methods are listed in Table~\ref{tab:DBLPEvaluationResults}, where it can be seen that the majority of the examined \AV methods yield useful recognition results with a maximum value of 0.792 in terms of \catOne. With the exception of the binary-intrinsic approach \coav, the remaining top performing methods belong to the binary-extrinsic category. This category of \AV methods has also been superior in the PAN-\AV competitions \cite{PANOverviewAV:2013,PANOverviewAV:2014,PANOverviewAV:2015}, where they outperformed binary-intrinsic and unary approaches three times in a row (2013--2015). 

The top performing approaches \bagnall, \coav and \veenmanNNCD deserve closer attention. All three are based on character-level language models that capture low-level features similar to character $n$-grams, which have been shown in numerous \AA and \AV studies (for instance, \cite{StamatatosRobustnessCharNGramsAA:2013,NealAVviaIsolationForests:2018}) to be highly effective and robust. In \cite{HalvaniARES:2017,BevendorffUnmasking:2019}, it has been shown that \bagnall and \coav were also the two top-performing approaches, where in \cite{HalvaniARES:2017} they were evaluated on the PAN-2015 \AV corpus \cite{PANOverviewAV:2015}, while in \cite{BevendorffUnmasking:2019} they were applied\footnote{Note that the implementation in \cite{BevendorffUnmasking:2019} differs from the one used in this paper.} on texts obtained from \e{Project Gutenberg}. Although both approaches perform similarly, they differ in the way how the decision criterion $\Threshold$ is determined. While \coav requires a training corpus to learn $\Threshold$, \bagnall assumes that the given test corpus (which provides the \textbf{impostors}) is balanced. Given this assumption, \bagnall first computes similarity scores for all verification problems in the corpus and then sets $\Threshold$ to the median of all similarities (cf. Figure~\ref{fig:PervCorpusShort}). Thus, from a machine learning perspective, there is some undue training on the test set. Moreover, the applicability of \bagnall in realistic scenarios is questionable, as a forensic case is \emph{not} part of a corpus where the \classY/\classN-distribution is \textbf{known beforehand}. 

Another interesting observation can be made regarding \coav, \veenmanNNCD and \occav. Although all three differ regarding their model category, they use the same underlying compression algorithm (PPMd) that is responsible for generating the language model. While the former two approaches perform similarly well, \occav achieves a poor \catOne score ($\approx0.5$). An obvious explanation for this is a wrongly calibrated threshold $\Threshold$, as can be seen from the confusion matrix, where almost all answers are \classN-predictions. Regarding the \veenmanNNCD approach, one should consider that $\Dunk$ is compared against $\Dknown$ as well as $n - 1$ impostors within a corpus comprised of $n$ verification problems. Therefore, a \classY-result is correct with relatively high certainty (\ie the method has high precision compared to other approaches with a similar \catOne score), as \veenmanNNCD decided that author $\A$ fits best to $\Dunk$ among $n$ candidates. In contrast to \bagnall, \veenmanNNCD only retrieves the impostors from the given corpus, but it does not exploit background knowledge about the distribution of problems in the corpus.

Overall, the results indicate that it is possible to recognize writing styles across large time spans. To gain more insights regarding the question which features led to the correct predictions, we inspected the \aveer method. Although the method achieved only average results, it benefits from the fact that it can be interpreted easily, as it relies on a simple distance function, a fixed threshold $\Threshold$ and predefined feature categories such as function words. Regarding the correctly recognized \classY-cases, we noticed that conjunctive adverbs such as \e{``hence''}, \e{``therefore''} or \e{``moreover''} contributed mostly to \aveer's correct predictions. However, a more in-depth analysis is required in future work to figure out whether the decisions of the remaining methods are also primarily affected by these features.

\subsubsection{The Effect of Topical Influence:}
In this experiment, we investigate the question if the writing style of authors can be recognized under the influence of topical bias. In real-world scenarios, the topic of the documents within a verification problem $\Problem$ is not always known beforehand, which can lead to a serious challenge regarding the recognition of the writing style. Imagine, for example, that $\Problem$ consists of a known and unknown document $\DA$ and $\Dunk$ that are written by the same author ($\A = \unknown$) while at the same time differ regarding their topic. In such a case, an \AV method that it focusing ``too much'' on the topic (for example on specific nouns or phrases) will likely predict a different authorship ($\A \neq \unknown$). On the other hand, when $\DA$ and $\Dunk$ match regarding their topic, while being written by different authors, a topically biased \AV method might erroneously predict $\A = \unknown$.  
\begin{table} 
	\begin{center}
		\begin{tabular}{lrrrrrrr} \toprule
			\textbf{AV Method} & \textbf{\catOne} & \textbf{\auc}  & \textbf{TP} & \textbf{FN} & \textbf{FP} & \textbf{TN}  &  \textbf{UP}  \\ \midrule			
			\koppelGI        & \textbf{0.533} & 0.521 & 22 & 8  & 20 & 10 & 0 \\
			\mocc            & 0.517          & {\color{gray} 0.537} & 8  & 22 & 7  & 23 & 0 \\
			\aveer           & 0.517          & 0.492 & 18 & 12 & 17 & 13 & 0 \\
			\veenmanNNCD     & 0.517          & {\color{gray} 0.561} & 13 & 17 & 12 & 18 & 0 \\
			\koppelUnmask    & 0.508          & \textbf{0.559} & 15 & 14 & 15 & 15 & 1 \\
			\stamatatosImpGI & 0.500          & 0.456  & 30 & 0  & 30 & 0  & 0 \\
			\occav           & 0.483          & {\color{gray} 0.428} & 0  & 30 & 1  & 29 & 0 \\
			\coav            & 0.450 & 0.532 & 18 & 12 & 21 & 9  & 0 \\
			\spatium         & 0.432 & 0.487 & 14 & 9  & 16 & 7  & 14 \\
			\bagnall         & 0.426 & 0.451 & 13 & 10 & 12 & 14 & 11 \\
			\noeckerDist     & 0.417 & 0.434 & 3  & 27 & 8  & 22 & 0 \\
			\glad            & 0.350 & 0.340 & 8  & 22 & 17 & 13 & 0 \\	 		
			\bottomrule			
		\end{tabular}
	\end{center}
	\caption{Evaluation results for the test corpus $\CorpusReddit$. \label{tab:RedditEvaluationResults}}
\end{table}
In the following we show to which extent these assumptions hold. As a data basis for this experiment, we used the $\CorpusReddit$ corpus introduced in Section~\ref{Corpus_Reddit}. The results regarding the \numberOfAVmethods \AV methods are given in Table~\ref{tab:RedditEvaluationResults}, where it can be seen that our assumptions hold. All examined \AV methods (with no exception) are fooled by the topical bias in the corpus. Here, the highest achieved results in terms of \catOne and \auc are very close to random guessing. A closer look at the confusion matrix outcomes reveals that some methods, for example \stamatatosImpGI and \occav, perform almost entirely inverse to each other, where the former predicts nothing but \classY and the latter nothing but \classN (except 1 \classY). Moreover, we can assume that the lower \catOne is, the stronger is the focus of the respective \AV method on the topic of the documents. Overall, the results of this experiment suggest that none of the examined \AV methods is robust against topical influence.

\subsubsection{The Effect of Limited Text Length:}
In our third experiment, we investigate the question how text lengths affect the results of the examined \AV methods. The motivation behind this experiment is based on the observation of Stamatatos et al. \cite{PANOverviewAV:2015} that text length is an important issue, which has not been thoroughly studied within \av research. To address this issue, we make use of the $\CorpusPeeJ$ corpus introduced in Section~\ref{Corpora_PeeJ}. The corpus is suitable for this purpose, as it comprises a large number of verification problems, where more than 90\% of all documents have sufficient text lengths ($\geq\,$2,000 characters). This allows a stepwise truncation and by this to analyze the effect between the text lengths and the recognition results. However, before considering this, we first focus on the results (shown in Table~\ref{tab:PeeJEvaluationResults}) after applying all \numberOfAVmethods \AV methods on the original test corpus. 
\begin{table} 	
	\begin{center}
		\begin{tabular}{lrrrrrrr} \toprule
			\textbf{AV Method} & \textbf{\catOne} & \textbf{\auc}  & \textbf{TP} & \textbf{FN} & \textbf{FP} & \textbf{TN}  &  \textbf{UP}  \\ \midrule	
			\veenmanNNCD     & \textbf{0.991} & {\color{gray} 0.999} & 325 & 5 & 1 & 329 & 0 \\			
			\bagnall	     & 0.973 & \textbf{0.995} & 319 & 9 & 9 & 319 & 4 \\			
			\glad            & 0.970 & \textbf{0.995} & 317 & 13 & 7 & 323 & 0 \\
			\aveer           & 0.924 & 0.975 & 298 & 32 & 18 & 312 & 0 \\
			\coav            & 0.923 & 0.980 & 303 & 27 & 24 & 306 & 0 \\
			\koppelUnmask    & 0.918 & 0.975 & 314 & 16 & 38 & 292 & 0 \\
			\spatium         & 0.892 & 0.971 & 312 & 10 & 53 & 235 & 50 \\
			\stamatatosImpGI & 0.886 & 0.974 & 322 & 8  & 67 & 263 & 0 \\
			\koppelGI        & 0.885 & 0.973 & 323 & 7  & 69 & 261 & 0 \\
			\mocc            & 0.853 & {\color{gray} 0.960} & 246 & 84 & 13 & 317 & 0 \\
			\noeckerDist     & 0.847 & {\color{gray} 0.913} & 282 & 48 & 53 & 277 & 0 \\
			\occav           & 0.798 & {\color{gray} 0.992} & 197 & 133 & 0 & 330 & 0 \\					
			\bottomrule			
		\end{tabular}
	\end{center}
	\caption{Evaluation results for the test corpus $\CorpusPeeJ$. \label{tab:PeeJEvaluationResults}}
\end{table}
As can be seen in Table~\ref{tab:PeeJEvaluationResults}, almost all approaches perform very well with \catOne scores up to 0.991. Although these results  are quite impressive, it should be noted that a large fraction of the documents comprises thousands of words. Thus, the methods can learn precise representations based on a large variety of features, which in turn enable a good determination of (dis)similarities between known/unknown documents. To investigate this issue in more detail, we constructed four versions of the test corpus and equalized the unknown document lengths to 250, 500, 1000, and 2000 characters. Then, we applied the top performing \AV methods with a \catOne value $> 0.9$ on the four corpora. Here, we reused the same models and hyperparameters (including the decision criteria $\Threshold$ and $\ThresholdModel$) that were determined on the training corpus. The intention behind this was to observe the robustness of the trained \AV models, given the fact that during training they were confronted with longer documents.  
\begin{figure}
  \centering
  \includegraphics[width=\linewidth]{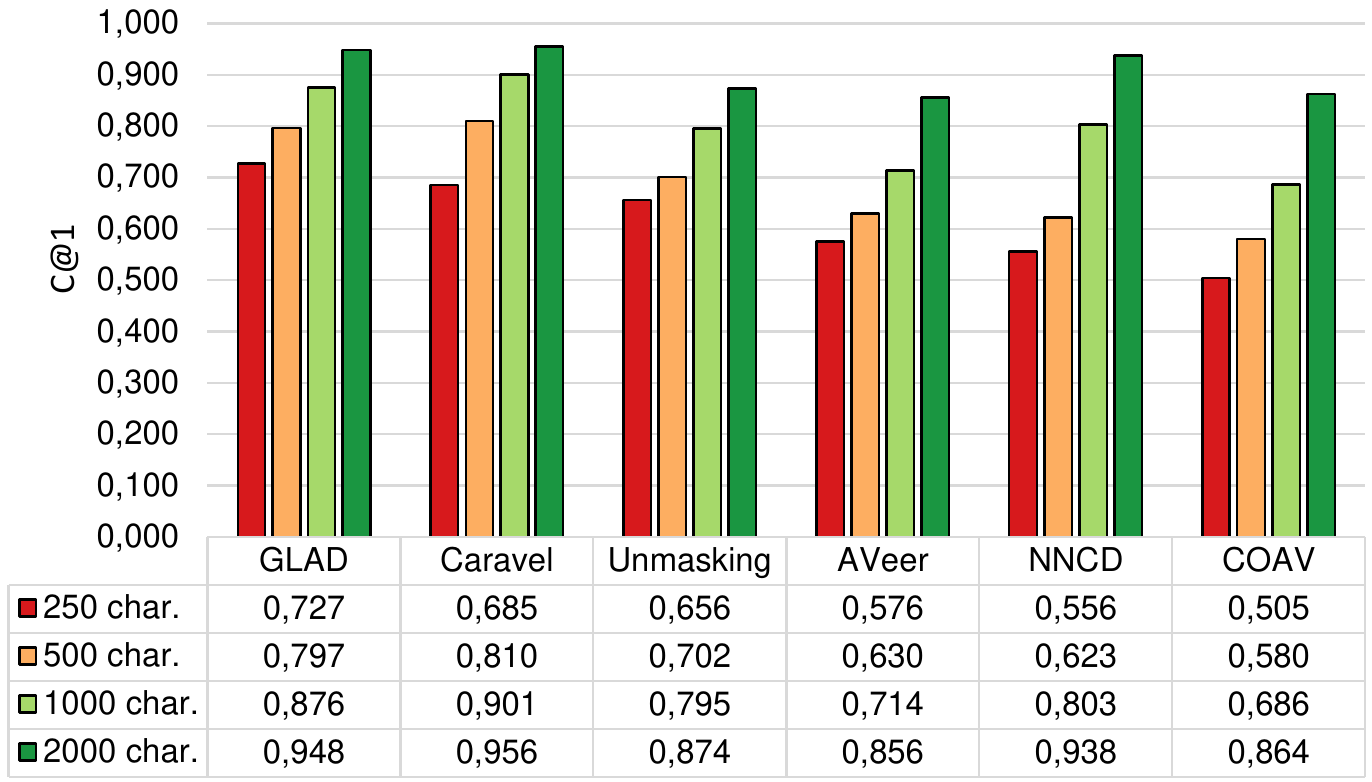}
  \caption{Evaluation results for the four versions of the test corpus $\CorpusPeeJ$ in terms of \catOne. \label{fig:PeeJEval_Textlengths}}
\end{figure}
The results are illustrated in Figure~\ref{fig:PeeJEval_Textlengths}, where it can be observed that \glad yields the most stable results across the four corpora versions, where even for the corpus with the 250 characters long unknown documents, it achieves a \catOne score of 0.727. Surprisingly, \koppelUnmask performs similarly well, despite of the fact that the method has been designed for longer texts \ie book chunks of \textbf{at least 500 words} \cite{KoppelAVOneClassClassification:2004}. Sanderson and Guenter also point out that the \koppelUnmask approach is less useful when dealing with relatively short texts \cite{SandersonUnmaskingAV:2006}. However, our results show a different picture, at least for this corpus.

\begin{figure*}
	\centering
	\includegraphics[width=0.3\textwidth]{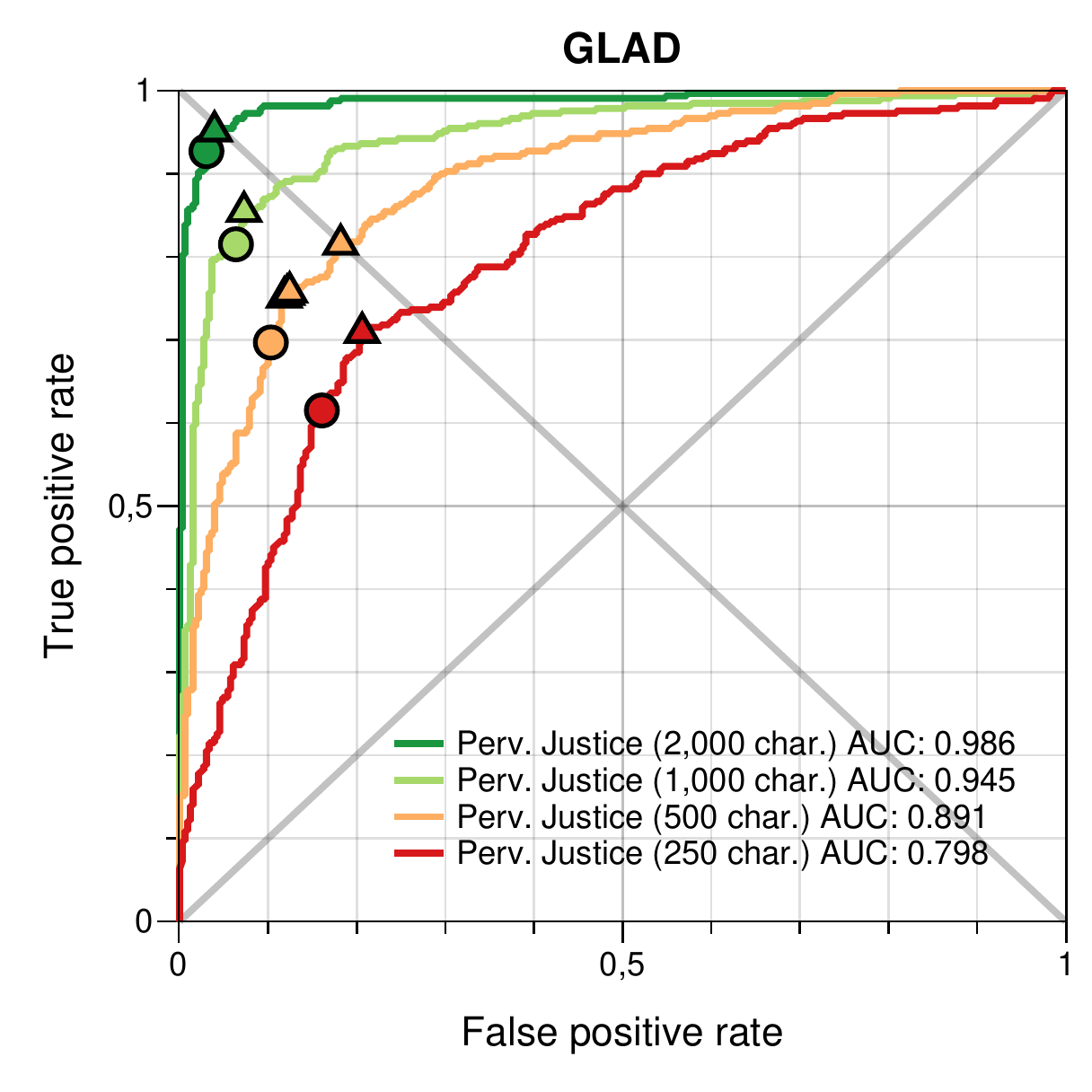}
	\includegraphics[width=0.3\textwidth]{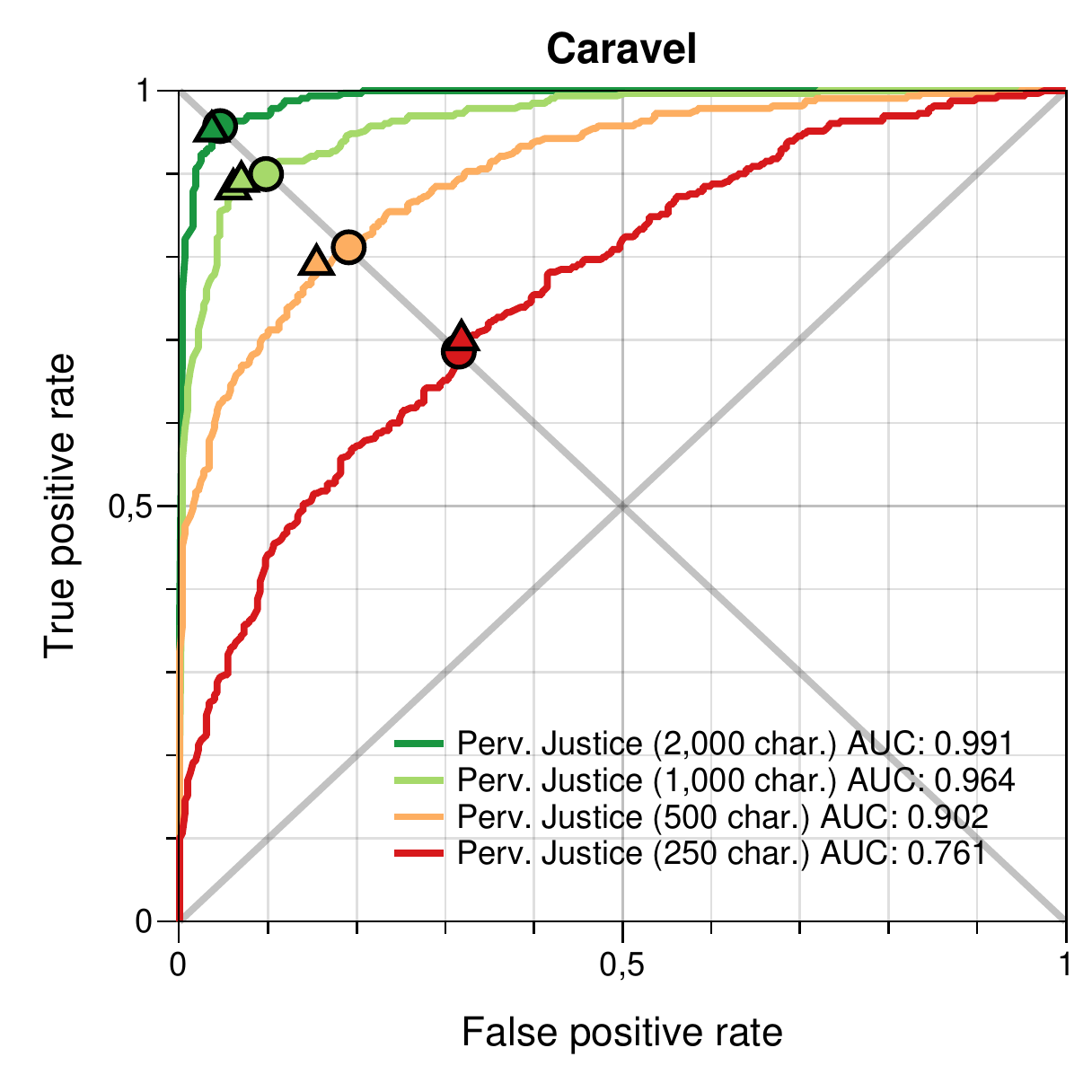}
	\includegraphics[width=0.3\textwidth]{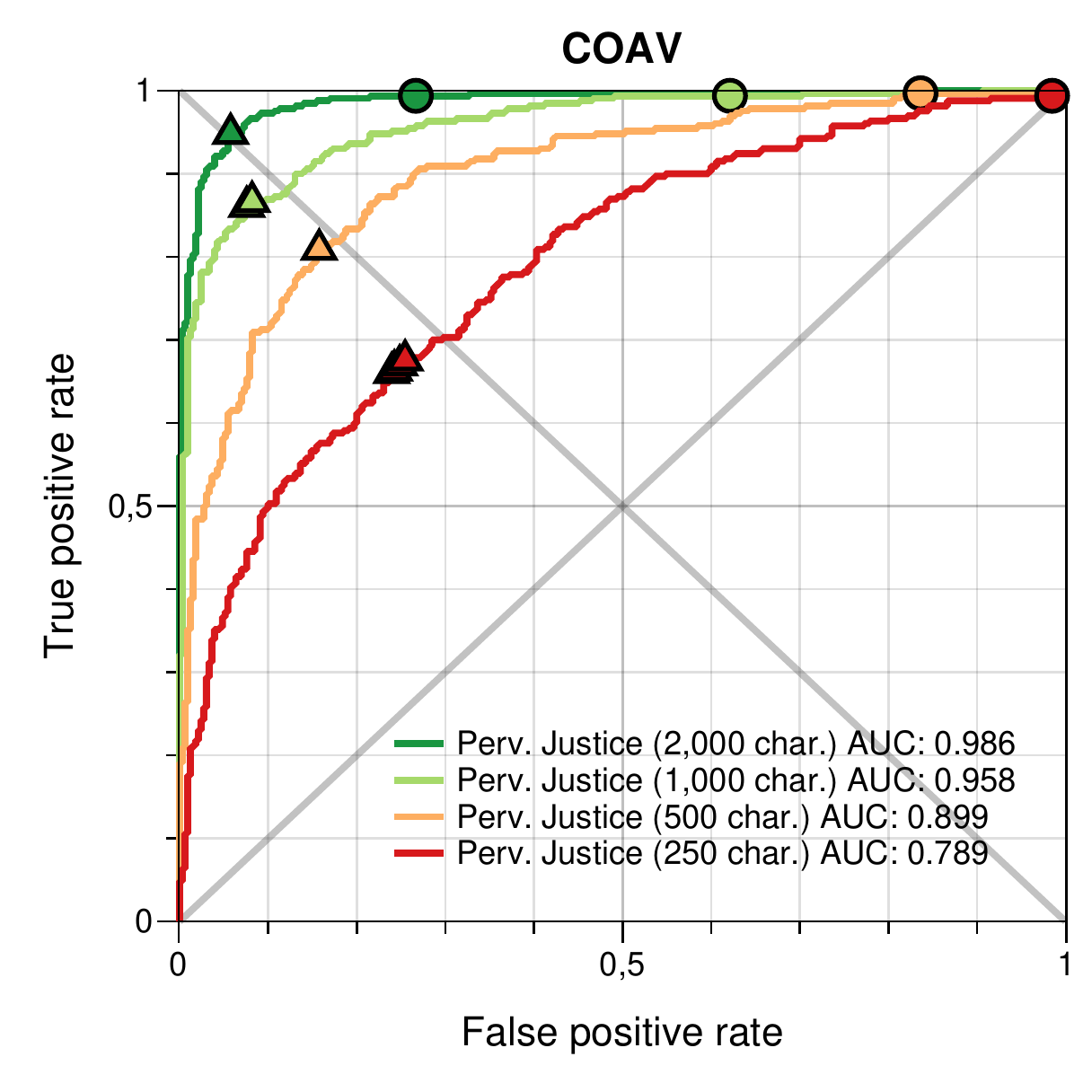}
	\caption{ROC curves for \glad, \bagnall and \coav (applied on the four corpora versions of $\CorpusPeeJ$). The circles and triangles depict the current and maximum achievable \catOne values on the corpus, respectively. Note that \bagnall's thresholds always lie along the EER-line. \label{fig:PervCorpusShort}}
\end{figure*}

One explanation of the resilience of \glad across the varying text lengths might be due to its decision model $\ThresholdModel$ (an SVM with a linear kernel) that withstands the absence of missing features caused by the truncation of the documents, in contrast to the distance-based approaches \aveer, \veenmanNNCD and \coav, where the decision criterion $\Threshold$ is reflected by a simple scalar. Table~\ref{tab:PeeJEvaluationResults250} lists the confusion matrix outcomes of the six \AV methods regarding the 250 characters version of $\CorpusPeeJ$. 
\begin{table}	
	\begin{center}
		\begin{tabular}{lrrrrr} \toprule
			\textbf{AV Method} & \textbf{TP} & \textbf{FN} & \textbf{FP} & \textbf{TN}  &  \textbf{Total (\classY/\classN/UP)}  \\ \midrule	
			\glad         & 203 & 127 &  53 & 277 & (256/404/0) \\			
			\bagnall      &	225	& 103 & 104 & 226 & (329/329/2) \\ 			
			\koppelUnmask & 158	& 169 &	 56	& 272 & (214/441/5) \\ 
			\aveer        & 56	& 274 &   6 & 324 & (62/598/0) \\  
			\veenmanNNCD  & 40  & 290 &   3 & 327 & (43/617/0) \\
			\coav         & 328	&   2 & 325 &   5 & (653/7/0) \\
			\bottomrule			
		\end{tabular}
	\end{center}
	\caption{Confusion matrix outcomes for the 250 characters version of the test corpus $\CorpusPeeJ$. \label{tab:PeeJEvaluationResults250}}
\end{table}
Here, it can be seen that the underlying SVM model of \glad and \koppelUnmask is able to regulate its \classY/\classN-predictions, in contrast to the three distance-based methods, where the majority of predictions fall either on the \classY- or on the \classN-side. To gain a better picture regarding the stability of the decision criteria $\Threshold$ and $\ThresholdModel$ of the methods, we decided to take a closer look on the ROC curves (cf. Figure~\ref{fig:PervCorpusShort}) generated by \glad, \bagnall and \coav for the four corpora versions, where a number of interesting observations can be made. When focusing on \auc, it turns out that all three methods perform very similar to each other, whereas big discrepancies between \glad and \coav can be observed regarding \catOne. When we consider the current and maximum achievable results (depicted by the circles and triangles, respectively) it becomes apparent that \glad's model behaves stable, while the one of \coav becomes increasingly vulnerable the more the documents are shortened. When looking at the ROC curve of \bagnall, it can be clearly seen that the actual and maximum achievable results are very close to each other. This is not surprising, due to the fact that \bagnall's threshold always lies at the median point of the ROC curve, provided that the given corpus is balanced. 
While inspecting the 250 characters long documents in more detail, we identified that they share similar vocabularies consisting of chat abbreviations such as \e{``lol''} (laughing out loud) or \e{``k''} (ok), smileys and specific obscene words. Therefore, we assume that the verification results of the examined methods are mainly caused by the similar vocabularies between the texts.

\section{Conclusion and Future Work} \label{Conclusions}
We highlighted the problem that underlying characteristics of \av approaches have not been paid much attention in the past research and that these affect the applicability of the methods in real forensic settings. Then, we proposed several properties that enable a better characterization and by this a better comparison between \AV methods. Among others, we explained that the performance measure \auc is meaningless in regard to \textbf{unary} or specific \textbf{non-optimizable} \AV methods, which involve a fixed decision criterion (for example, \veenmanNNCD). Additionally, we mentioned that determinism must be fulfilled such that an \AV method can be rated as reliable. Moreover, we clarified a number of misunderstandings in previous research works and proposed three clear criteria that allow to classify the model category of an \AV method, which in turn influences its design and the way how it should be evaluated. In regard to binary-extrinsic \AV approaches, we explained which challenges exist and how they affect their applicability.
 
In an experimental setup, we applied \numberOfAVmethods existing \AV methods on three self-compiled corpora, where the intention behind each corpus was to focus on a different aspect of the methods applicability. Our findings regarding the examined approaches can be summarized as follows: 
Despite of the good performance of the five \AV methods \koppelGI, \stamatatosImpGI, \koppelUnmask, \bagnall and \spatium, none of them can be truly considered as reliable and therefore applicable in real forensic cases. The reason for this is not only the non-deterministic behavior of the methods but also their dependence (excepting \koppelUnmask) on an impostor corpus. Here, it must be guaranteed that the true author is not among the candidates, but also that the impostor documents are suitable such that the \AV task not inadvertently degenerates from style to topic classification. In particular, the applicability of the \bagnall approach remains highly questionable, as it requires a corpus where the information regarding \classY/\classN-distribution is known beforehand in order to set the threshold. In regard to the two examined unary \AV approaches \mocc and \occav, we observed that these perform poorly on all three corpora in comparison to the binary-intrinsic and binary-extrinsic methods. Most likely, this is caused by the wrong threshold setting, as both tend to generate more \classN-predictions. From the remaining approaches, \glad and \coav seem to be a good choice for realistic scenarios. However, the former has been shown to be more robust in regard to varying text lengths given a fixed model, while the latter requires a retraining of the model (note that both performed almost equal in terms of \auc). Our hypothesis, which we leave open for future work, is that \AV methods relying on a complex model $\ThresholdModel$ are more robust than methods based on a scalar-threshold $\Threshold$. Lastly, we wish to underline that all examined approaches failed in the cross-topic experiment. One possibility to counteract this is to apply text distortion techniques (for instance, \cite{StamatatosTextDistortion:2017}) in order to control the topic influence in the documents. 

As one next step, we will compile additional and larger corpora to investigate the question whether the evaluation results of this paper hold more generally. Furthermore, we will address the important question how the results of \AV methods can be \emph{interpreted} in a more systematic manner, which will further influence the practicability of \AV methods besides the proposed properties. 

\begin{acks}
This work was supported by the German Federal Ministry of Education and Research (BMBF) under the project "DORIAN" (Scrutinise and thwart disinformation).
\end{acks}

\bibliographystyle{ACM-Reference-Format}
\bibliography{Bibliography}


\begin{thebibliography}{42}


\ifx \showCODEN    \undefined \def \showCODEN     #1{\unskip}     \fi
\ifx \showDOI      \undefined \def \showDOI       #1{#1}\fi
\ifx \showISBNx    \undefined \def \showISBNx     #1{\unskip}     \fi
\ifx \showISBNxiii \undefined \def \showISBNxiii  #1{\unskip}     \fi
\ifx \showISSN     \undefined \def \showISSN      #1{\unskip}     \fi
\ifx \showLCCN     \undefined \def \showLCCN      #1{\unskip}     \fi
\ifx \shownote     \undefined \def \shownote      #1{#1}          \fi
\ifx \showarticletitle \undefined \def \showarticletitle #1{#1}   \fi
\ifx \showURL      \undefined \def \showURL       {\relax}        \fi
\providecommand\bibfield[2]{#2}
\providecommand\bibinfo[2]{#2}
\providecommand\natexlab[1]{#1}
\providecommand\showeprint[2][]{arXiv:#2}

\bibitem[\protect\citeauthoryear{Azarbonyad, Dehghani, Marx, and
  Kamps}{Azarbonyad et~al\mbox{.}}{2015}]%
        {AzarbonyadTimeAwareAA:2015}
\bibfield{author}{\bibinfo{person}{Hosein Azarbonyad}, \bibinfo{person}{Mostafa
  Dehghani}, \bibinfo{person}{Maarten Marx}, {and} \bibinfo{person}{Jaap
  Kamps}.} \bibinfo{year}{2015}\natexlab{}.
\newblock \showarticletitle{{Time-Aware Authorship Attribution for Short Text
  Streams}}. In \bibinfo{booktitle}{\emph{Proceedings of the 38th International
  ACM SIGIR Conference on Research and Development in Information Retrieval}}
  \emph{(\bibinfo{series}{SIGIR '15})}. \bibinfo{publisher}{ACM},
  \bibinfo{address}{New York, NY, USA}, \bibinfo{pages}{727--730}.
\newblock
\showISBNx{978-1-4503-3621-5}


\bibitem[\protect\citeauthoryear{Bagnall}{Bagnall}{2015}]%
        {BagnallRNN:2015}
\bibfield{author}{\bibinfo{person}{Douglas Bagnall}.}
  \bibinfo{year}{2015}\natexlab{}.
\newblock \showarticletitle{{Author Identification Using Multi-headed Recurrent
  Neural Networks}}. In \bibinfo{booktitle}{\emph{Working Notes of {CLEF} 2015
  - Conference and Labs of the Evaluation forum, Toulouse, France, September
  8-11, 2015.}}
\newblock


\bibitem[\protect\citeauthoryear{Bevendorff, Stein, Hagen, and
  Potthast}{Bevendorff et~al\mbox{.}}{2019}]%
        {BevendorffUnmasking:2019}
\bibfield{author}{\bibinfo{person}{Janek Bevendorff}, \bibinfo{person}{Benno
  Stein}, \bibinfo{person}{Matthias Hagen}, {and} \bibinfo{person}{Martin
  Potthast}.} \bibinfo{year}{2019}\natexlab{}.
\newblock \showarticletitle{{Generalizing Unmasking for Short Texts}}. In
  \bibinfo{booktitle}{\emph{Proceedings of the 2019 Conference of the North
  {A}merican Chapter of the Association for Computational Linguistics: Human
  Language Technologies, Volume 1 (Long and Short Papers)}}.
  \bibinfo{publisher}{Association for Computational Linguistics},
  \bibinfo{address}{Minneapolis, Minnesota}, \bibinfo{pages}{654--659}.
\newblock


\bibitem[\protect\citeauthoryear{Bollen}{Bollen}{1989}]%
        {BollenReliability:1989}
\bibfield{author}{\bibinfo{person}{Kenneth~A. Bollen}.}
  \bibinfo{year}{1989}\natexlab{}.
\newblock \bibinfo{booktitle}{\emph{{Structural Equations with Latent
  Variables}}}.
\newblock \bibinfo{publisher}{Wiley}.
\newblock
\showISBNx{9780471011712}
\showLCCN{lc88027272}


\bibitem[\protect\citeauthoryear{Boukhaled and Ganascia}{Boukhaled and
  Ganascia}{2014}]%
        {BoukhaledProbabilisticAV:2014}
\bibfield{author}{\bibinfo{person}{Mohamed~Amine Boukhaled} {and}
  \bibinfo{person}{Jean-Gabriel Ganascia}.} \bibinfo{year}{2014}\natexlab{}.
\newblock \bibinfo{booktitle}{\emph{{Probabilistic Anomaly Detection Method for
  Authorship Verification}}}.
\newblock \bibinfo{publisher}{Springer International Publishing},
  \bibinfo{address}{Cham}, \bibinfo{pages}{211--219}.
\newblock
\showISBNx{978-3-319-11397-5}


\bibitem[\protect\citeauthoryear{Castro~Castro, Adame~Arcia, Pelaez~Brioso, and
  Mu{\~{n}}oz~Guillena}{Castro~Castro et~al\mbox{.}}{2015}]%
        {CastroAVAverageSimilarity:2015}
\bibfield{author}{\bibinfo{person}{Daniel Castro~Castro},
  \bibinfo{person}{Yaritza Adame~Arcia}, \bibinfo{person}{Mar{\'i}a
  Pelaez~Brioso}, {and} \bibinfo{person}{Rafael Mu{\~{n}}oz~Guillena}.}
  \bibinfo{year}{2015}\natexlab{}.
\newblock \showarticletitle{{Authorship Verification, Average Similarity
  Analysis}}. In \bibinfo{booktitle}{\emph{Proceedings of the International
  Conference Recent Advances in Natural Language Processing}}.
  \bibinfo{publisher}{INCOMA Ltd. Shoumen, BULGARIA}, \bibinfo{pages}{84--90}.
\newblock


\bibitem[\protect\citeauthoryear{Gr{\"{o}}ndahl and Asokan}{Gr{\"{o}}ndahl and
  Asokan}{2019}]%
        {GrondahlTextanalysisAdversarial:2019}
\bibfield{author}{\bibinfo{person}{Tommi Gr{\"{o}}ndahl} {and}
  \bibinfo{person}{N. Asokan}.} \bibinfo{year}{2019}\natexlab{}.
\newblock \showarticletitle{{Text Analysis in Adversarial Settings: Does
  Deception Leave a Stylistic Trace?}}
\newblock \bibinfo{journal}{\emph{CoRR}}  \bibinfo{volume}{abs/1902.08939}
  (\bibinfo{year}{2019}).
\newblock
\showeprint[arxiv]{1902.08939}


\bibitem[\protect\citeauthoryear{Halvani, Graner, and Vogel}{Halvani
  et~al\mbox{.}}{2018}]%
        {HalvaniOCCAV:2018}
\bibfield{author}{\bibinfo{person}{Oren Halvani}, \bibinfo{person}{Lukas
  Graner}, {and} \bibinfo{person}{Inna Vogel}.}
  \bibinfo{year}{2018}\natexlab{}.
\newblock \showarticletitle{{Authorship Verification in the Absence of Explicit
  Features and Thresholds}}. In \bibinfo{booktitle}{\emph{Advances in
  Information Retrieval}}, \bibfield{editor}{\bibinfo{person}{Gabriella Pasi},
  \bibinfo{person}{Benjamin Piwowarski}, \bibinfo{person}{Leif Azzopardi},
  {and} \bibinfo{person}{Allan Hanbury}} (Eds.). \bibinfo{publisher}{Springer
  International Publishing}, \bibinfo{pages}{454--465}.
\newblock
\showISBNx{978-3-319-76941-7}


\bibitem[\protect\citeauthoryear{Halvani and Steinebach}{Halvani and
  Steinebach}{2014}]%
        {HalvaniARES:2014}
\bibfield{author}{\bibinfo{person}{Oren Halvani} {and} \bibinfo{person}{Martin
  Steinebach}.} \bibinfo{year}{2014}\natexlab{}.
\newblock \showarticletitle{{An Efficient Intrinsic Authorship Verification
  Scheme Based on Ensemble Learning}}. In \bibinfo{booktitle}{\emph{Ninth
  International Conference on Availability, Reliability and Security, {ARES}
  2014, Fribourg, Switzerland, September 8-12, 2014}}.
  \bibinfo{address}{Washington, DC, USA}, \bibinfo{pages}{571--578}.
\newblock
\showISBNx{978-1-4799-4223-7}


\bibitem[\protect\citeauthoryear{Halvani, Winter, and Graner}{Halvani
  et~al\mbox{.}}{2017}]%
        {HalvaniARES:2017}
\bibfield{author}{\bibinfo{person}{Oren Halvani}, \bibinfo{person}{Christian
  Winter}, {and} \bibinfo{person}{Lukas Graner}.}
  \bibinfo{year}{2017}\natexlab{}.
\newblock \showarticletitle{{On the Usefulness of Compression Models for
  Authorship Verification}}. In \bibinfo{booktitle}{\emph{Proceedings of the
  12th International Conference on Availability, Reliability and Security}}
  \emph{(\bibinfo{series}{ARES '17})}. \bibinfo{publisher}{ACM},
  \bibinfo{address}{New York, NY, USA}, Article \bibinfo{articleno}{54},
  \bibinfo{numpages}{10}~pages.
\newblock
\showISBNx{978-1-4503-5257-4}


\bibitem[\protect\citeauthoryear{Halvani, Winter, and Pflug}{Halvani
  et~al\mbox{.}}{2016}]%
        {HalvaniDFRWS:2016}
\bibfield{author}{\bibinfo{person}{Oren Halvani}, \bibinfo{person}{Christian
  Winter}, {and} \bibinfo{person}{Anika Pflug}.}
  \bibinfo{year}{2016}\natexlab{}.
\newblock \showarticletitle{{Authorship Verification for Different Languages,
  Genres and Topics}}.
\newblock \bibinfo{journal}{\emph{Digit. Investig.}} \bibinfo{volume}{16},
  \bibinfo{number}{S} (\bibinfo{date}{March} \bibinfo{year}{2016}),
  \bibinfo{pages}{S33--S43}.
\newblock
\showISSN{1742-2876}


\bibitem[\protect\citeauthoryear{Hern{\'{a}}ndez, Casillas, Ledesma, Pineda,
  and Ru{\'{\i}}z}{Hern{\'{a}}ndez et~al\mbox{.}}{2015}]%
        {HernandezHomotopyAV:2015}
\bibfield{author}{\bibinfo{person}{Josu{\'{e}} Gerardo~Guti{\'{e}}rrez
  Hern{\'{a}}ndez}, \bibinfo{person}{Jos{\'{e}} Casillas},
  \bibinfo{person}{Paola Ledesma}, \bibinfo{person}{Gibran~Fuentes Pineda},
  {and} \bibinfo{person}{Iv{\'{a}}n Vladimir~Meza Ru{\'{\i}}z}.}
  \bibinfo{year}{2015}\natexlab{}.
\newblock \showarticletitle{{Homotopy Based Classification for Author
  Verification Task: Notebook for {PAN} at {CLEF} 2015}}. In
  \bibinfo{booktitle}{\emph{Working Notes of {CLEF} 2015 - Conference and Labs
  of the Evaluation forum, Toulouse, France, September 8-11, 2015.}}
\newblock


\bibitem[\protect\citeauthoryear{Hern{\'{a}}ndez{-}Casta{\~{n}}eda and
  Calvo}{Hern{\'{a}}ndez{-}Casta{\~{n}}eda and Calvo}{2017}]%
        {CastanedaAVviaLDA:2017}
\bibfield{author}{\bibinfo{person}{{\'{A}}ngel
  Hern{\'{a}}ndez{-}Casta{\~{n}}eda} {and} \bibinfo{person}{Hiram Calvo}.}
  \bibinfo{year}{2017}\natexlab{}.
\newblock \showarticletitle{{Author Verification Using a Semantic Space
  Model}}.
\newblock \bibinfo{journal}{\emph{Computaci{\'{o}}n y Sistemas}}
  \bibinfo{volume}{21}, \bibinfo{number}{2} (\bibinfo{year}{2017}).
\newblock


\bibitem[\protect\citeauthoryear{Holmes}{Holmes}{1998}]%
        {HolmesEvolutionStylometryHumanities:1998}
\bibfield{author}{\bibinfo{person}{David~I. Holmes}.}
  \bibinfo{year}{1998}\natexlab{}.
\newblock \showarticletitle{{The Evolution of Stylometry in Humanities
  Scholarship}}.
\newblock \bibinfo{journal}{\emph{Literary and Linguistic Computing}}
  \bibinfo{volume}{13}, \bibinfo{number}{3} (\bibinfo{year}{1998}),
  \bibinfo{pages}{111--117}.
\newblock


\bibitem[\protect\citeauthoryear{H{\"u}rlimann, Weck, von~den Berg,
  {\v{S}}uster, and Nissim}{H{\"u}rlimann et~al\mbox{.}}{2015}]%
        {GLAD:2015}
\bibfield{author}{\bibinfo{person}{Manuela H{\"u}rlimann},
  \bibinfo{person}{Benno Weck}, \bibinfo{person}{Esther von~den Berg},
  \bibinfo{person}{Simon {\v{S}}uster}, {and} \bibinfo{person}{Malvina
  Nissim}.} \bibinfo{year}{2015}\natexlab{}.
\newblock \showarticletitle{{GLAD: Groningen Lightweight Authorship
  Detection}}. In \bibinfo{booktitle}{\emph{Working Notes of {CLEF} 2015 --
  Conference and Labs of the Evaluation forum, Toulouse, France, September
  8--11, 2015}}. 12.
\newblock


\bibitem[\protect\citeauthoryear{Jankowska, Keselj, and Milios}{Jankowska
  et~al\mbox{.}}{2013}]%
        {JankowskaCNGAV:2013}
\bibfield{author}{\bibinfo{person}{Magdalena Jankowska}, \bibinfo{person}{Vlado
  Keselj}, {and} \bibinfo{person}{Evangelos~E. Milios}.}
  \bibinfo{year}{2013}\natexlab{}.
\newblock \showarticletitle{{Proximity Based One-class Classification with
  Common N-Gram Dissimilarity for Authorship Verification Task Notebook for
  {PAN} at {CLEF} 2013}}. In \bibinfo{booktitle}{\emph{Working Notes for {CLEF}
  2013 Conference , Valencia, Spain, September 23-26, 2013.}}
\newblock


\bibitem[\protect\citeauthoryear{Jankowska, Milios, and Keselj}{Jankowska
  et~al\mbox{.}}{2014}]%
        {JankowskaAVviaCNG:2014}
\bibfield{author}{\bibinfo{person}{Magdalena Jankowska},
  \bibinfo{person}{Evangelos~E. Milios}, {and} \bibinfo{person}{Vlado Keselj}.}
  \bibinfo{year}{2014}\natexlab{}.
\newblock \showarticletitle{{Author Verification Using Common N-Gram Profiles
  of Text Documents}}. In \bibinfo{booktitle}{\emph{{COLING} 2014, 25th
  International Conference on Computational Linguistics, Proceedings of the
  Conference: Technical Papers, August 23-29, 2014, Dublin, Ireland}},
  \bibfield{editor}{\bibinfo{person}{Jan Hajic} {and} \bibinfo{person}{Junichi
  Tsujii}} (Eds.). \bibinfo{publisher}{{ACL}}, \bibinfo{pages}{387--397}.
\newblock
\showISBNx{978-1-941643-26-6}


\bibitem[\protect\citeauthoryear{Jr and Ryan}{Jr and Ryan}{2012}]%
        {NoeckerDistractorlessAV:2012}
\bibfield{author}{\bibinfo{person}{John~Noecker Jr} {and}
  \bibinfo{person}{Michael Ryan}.} \bibinfo{year}{2012}\natexlab{}.
\newblock \showarticletitle{{Distractorless Authorship Verification}}. In
  \bibinfo{booktitle}{\emph{Proceedings of the Eight International Conference
  on Language Resources and Evaluation (LREC'12)}} (23-25),
  \bibfield{editor}{\bibinfo{person}{Nicoletta Calzolari~(Conference Chair)},
  \bibinfo{person}{Khalid Choukri}, \bibinfo{person}{Thierry Declerck},
  \bibinfo{person}{Mehmet~Uğur Doğan}, \bibinfo{person}{Bente Maegaard},
  \bibinfo{person}{Joseph Mariani}, \bibinfo{person}{Asuncion Moreno},
  \bibinfo{person}{Jan Odijk}, {and} \bibinfo{person}{Stelios Piperidis}}
  (Eds.). \bibinfo{publisher}{European Language Resources Association (ELRA)},
  \bibinfo{address}{Istanbul, Turkey}.
\newblock
\showISBNx{978-2-9517408-7-7}


\bibitem[\protect\citeauthoryear{Juola and Stamatatos}{Juola and
  Stamatatos}{2013}]%
        {PANOverviewAV:2013}
\bibfield{author}{\bibinfo{person}{Patrick Juola} {and}
  \bibinfo{person}{Efstathios Stamatatos}.} \bibinfo{year}{2013}\natexlab{}.
\newblock \showarticletitle{{Overview of the Author Identification Task at
  {PAN} 2013}}. In \bibinfo{booktitle}{\emph{Working Notes for {CLEF} 2013
  Conference, Valencia, Spain, September 23-26, 2013}}. 20.
\newblock


\bibitem[\protect\citeauthoryear{Khonji and Iraqi}{Khonji and Iraqi}{2014}]%
        {KhonjiIraqiAV:2014}
\bibfield{author}{\bibinfo{person}{Mahmoud Khonji} {and}
  \bibinfo{person}{Youssef Iraqi}.} \bibinfo{year}{2014}\natexlab{}.
\newblock \showarticletitle{{A Slightly-Modified GI-Based Author-Verifier with
  Lots of Features {(ASGALF)}}}. In \bibinfo{booktitle}{\emph{Working Notes for
  {CLEF} 2014 Conference, Sheffield, UK, September 15-18, 2014.}}
  \bibinfo{pages}{977--983}.
\newblock


\bibitem[\protect\citeauthoryear{Kocher and Savoy}{Kocher and Savoy}{2015}]%
        {KocherPANSpatium:2015}
\bibfield{author}{\bibinfo{person}{Mirco Kocher} {and} \bibinfo{person}{Jacques
  Savoy}.} \bibinfo{year}{2015}\natexlab{}.
\newblock \showarticletitle{UniNE at {CLEF} 2015 Author Identification:
  Notebook for {PAN} at {CLEF} 2015}. In \bibinfo{booktitle}{\emph{{CLEF}
  (Working Notes)}} \emph{(\bibinfo{series}{{CEUR} Workshop Proceedings})},
  Vol.~\bibinfo{volume}{1391}. \bibinfo{publisher}{CEUR-WS.org}.
\newblock


\bibitem[\protect\citeauthoryear{Koppel and Schler}{Koppel and Schler}{2004}]%
        {KoppelAVOneClassClassification:2004}
\bibfield{author}{\bibinfo{person}{Moshe Koppel} {and}
  \bibinfo{person}{Jonathan Schler}.} \bibinfo{year}{2004}\natexlab{}.
\newblock \showarticletitle{{Authorship Verification as a One-Class
  Classification Problem}}. In \bibinfo{booktitle}{\emph{Machine Learning,
  Proceedings of the Twenty-first International Conference {(ICML} 2004),
  Banff, Alberta, Canada, July 4-8, 2004}} \emph{(\bibinfo{series}{{ACM}
  International Conference Proceeding Series})},
  \bibfield{editor}{\bibinfo{person}{Carla~E. Brodley}} (Ed.),
  Vol.~\bibinfo{volume}{69}. \bibinfo{publisher}{{ACM}}.
\newblock


\bibitem[\protect\citeauthoryear{Koppel and Winter}{Koppel and Winter}{2014}]%
        {KoppelWinter2DocsBy1:2014}
\bibfield{author}{\bibinfo{person}{Moshe Koppel} {and} \bibinfo{person}{Yaron
  Winter}.} \bibinfo{year}{2014}\natexlab{}.
\newblock \showarticletitle{{Determining if Two Documents are Written by the
  Same Author}}.
\newblock \bibinfo{journal}{\emph{{JASIST}}} \bibinfo{volume}{65},
  \bibinfo{number}{1} (\bibinfo{year}{2014}), \bibinfo{pages}{178--187}.
\newblock


\bibitem[\protect\citeauthoryear{Neal, Sundararajan, and Woodard}{Neal
  et~al\mbox{.}}{2018}]%
        {NealAVviaIsolationForests:2018}
\bibfield{author}{\bibinfo{person}{Tempestt~J. Neal},
  \bibinfo{person}{Kalaivani Sundararajan}, {and} \bibinfo{person}{Damon~L.
  Woodard}.} \bibinfo{year}{2018}\natexlab{}.
\newblock \showarticletitle{{Exploiting Linguistic Style as a Cognitive
  Biometric for Continuous Verification}}. In \bibinfo{booktitle}{\emph{2018
  International Conference on Biometrics, {ICB} 2018, Gold Coast, Australia,
  February 20-23, 2018}}. \bibinfo{publisher}{{IEEE}},
  \bibinfo{pages}{270--276}.
\newblock
\showISBNx{978-1-5386-4285-6}


\bibitem[\protect\citeauthoryear{Olsson}{Olsson}{2008}]%
        {ForensicLinguisticsBookOlsson:2008}
\bibfield{author}{\bibinfo{person}{J. Olsson}.}
  \bibinfo{year}{2008}\natexlab{}.
\newblock \bibinfo{booktitle}{\emph{{Forensic Linguistics: Second Edition: An
  Introduction To Language, Crime and the Law}}}.
\newblock \bibinfo{publisher}{Bloomsbury Academic}.
\newblock
\showISBNx{9780826493088}
\showLCCN{2008277530}


\bibitem[\protect\citeauthoryear{Potha and Stamatatos}{Potha and
  Stamatatos}{2014}]%
        {StamatatosProfileCNG:2014}
\bibfield{author}{\bibinfo{person}{Nektaria Potha} {and}
  \bibinfo{person}{Efstathios Stamatatos}.} \bibinfo{year}{2014}\natexlab{}.
\newblock \showarticletitle{{A Profile-Based Method for Authorship
  Verification}}. In \bibinfo{booktitle}{\emph{Artificial Intelligence: Methods
  and Applications: 8th Hellenic Conference on AI, SETN 2014, Ioannina, Greece,
  May 15--17, 2014. Proceedings}}. \bibinfo{publisher}{Springer International
  Publishing}, \bibinfo{pages}{313--326}.
\newblock
\showISBNx{978-3-319-07064-3}


\bibitem[\protect\citeauthoryear{Potha and Stamatatos}{Potha and
  Stamatatos}{2017}]%
        {StamatatosPothaImprovedIM:2017}
\bibfield{author}{\bibinfo{person}{Nektaria Potha} {and}
  \bibinfo{person}{Efstathios Stamatatos}.} \bibinfo{year}{2017}\natexlab{}.
\newblock \showarticletitle{{An Improved Impostors Method for Authorship
  Verification}}. In \bibinfo{booktitle}{\emph{Experimental {IR} Meets
  Multilinguality, Multimodality, and Interaction - 8th International
  Conference of the {CLEF} Association, {CLEF} 2017, Dublin, Ireland, September
  11-14, 2017, Proceedings}}. \bibinfo{pages}{138--144}.
\newblock


\bibitem[\protect\citeauthoryear{Potha and Stamatatos}{Potha and
  Stamatatos}{2018}]%
        {PothaStamatatosTopicAV:2018}
\bibfield{author}{\bibinfo{person}{Nektaria Potha} {and}
  \bibinfo{person}{Efstathios Stamatatos}.} \bibinfo{year}{2018}\natexlab{}.
\newblock \showarticletitle{{Intrinsic Author Verification Using Topic
  Modeling}}. In \bibinfo{booktitle}{\emph{Proceedings of the 10th Hellenic
  Conference on Artificial Intelligence, {SETN} 2018, Patras, Greece, July
  09-12, 2018}}. \bibinfo{publisher}{{ACM}}, \bibinfo{pages}{20:1--20:7}.
\newblock


\bibitem[\protect\citeauthoryear{Potthast, Hagen, and Stein}{Potthast
  et~al\mbox{.}}{2016}]%
        {PANOverviewAO:2016}
\bibfield{author}{\bibinfo{person}{Martin Potthast}, \bibinfo{person}{Matthias
  Hagen}, {and} \bibinfo{person}{Benno Stein}.}
  \bibinfo{year}{2016}\natexlab{}.
\newblock \showarticletitle{{Author Obfuscation: Attacking the State of the Art
  in Authorship Verification}}. In \bibinfo{booktitle}{\emph{Working Notes
  Papers of the CLEF 2016 Evaluation Labs}} \emph{(\bibinfo{series}{CEUR
  Workshop Proceedings})}, Vol.~\bibinfo{volume}{1609}.
  \bibinfo{publisher}{CLEF and CEUR-WS.org}, \bibinfo{pages}{716--749}.
\newblock
\showISSN{1613-0073}


\bibitem[\protect\citeauthoryear{Potthast, Rosso, Stamatatos, and
  Stein}{Potthast et~al\mbox{.}}{2019}]%
        {DigitalTextForensicsECIR:2019}
\bibfield{author}{\bibinfo{person}{Martin Potthast}, \bibinfo{person}{Paolo
  Rosso}, \bibinfo{person}{Efstathios Stamatatos}, {and} \bibinfo{person}{Benno
  Stein}.} \bibinfo{year}{2019}\natexlab{}.
\newblock \showarticletitle{A Decade of Shared Tasks in Digital Text Forensics
  at PAN}. In \bibinfo{booktitle}{\emph{Advances in Information Retrieval}},
  \bibfield{editor}{\bibinfo{person}{Leif Azzopardi}, \bibinfo{person}{Benno
  Stein}, \bibinfo{person}{Norbert Fuhr}, \bibinfo{person}{Philipp Mayr},
  \bibinfo{person}{Claudia Hauff}, {and} \bibinfo{person}{Djoerd Hiemstra}}
  (Eds.). \bibinfo{publisher}{Springer International Publishing},
  \bibinfo{address}{Cham}, \bibinfo{pages}{291--300}.
\newblock
\showISBNx{978-3-030-15719-7}


\bibitem[\protect\citeauthoryear{Rodionova, Oliveri, and Pomerantsev}{Rodionova
  et~al\mbox{.}}{2016}]%
        {RodionovaOCC:2016}
\bibfield{author}{\bibinfo{person}{Oxana~Ye. Rodionova}, \bibinfo{person}{Paolo
  Oliveri}, {and} \bibinfo{person}{Alexey~L. Pomerantsev}.}
  \bibinfo{year}{2016}\natexlab{}.
\newblock \showarticletitle{{Rigorous and Compliant Approaches to One-Class
  Classification}}.
\newblock \bibinfo{journal}{\emph{Chemometrics and Intelligent Laboratory
  Systems}}  \bibinfo{volume}{159} (\bibinfo{year}{2016}), \bibinfo{pages}{89
  -- 96}.
\newblock
\showISSN{0169-7439}


\bibitem[\protect\citeauthoryear{Sanderson and Guenter}{Sanderson and
  Guenter}{2006}]%
        {SandersonUnmaskingAV:2006}
\bibfield{author}{\bibinfo{person}{Conrad Sanderson} {and}
  \bibinfo{person}{Simon Guenter}.} \bibinfo{year}{2006}\natexlab{}.
\newblock \showarticletitle{{Short Text Authorship Attribution via Sequence
  Kernels, Markov Chains and Author Unmasking: An Investigation}}. In
  \bibinfo{booktitle}{\emph{Proceedings of the 2006 Conference on Empirical
  Methods in Natural Language Processing}} \emph{(\bibinfo{series}{EMNLP
  '06})}. \bibinfo{publisher}{Association for Computational Linguistics},
  \bibinfo{address}{Stroudsburg, PA, USA}, \bibinfo{pages}{482--491}.
\newblock
\showISBNx{1-932432-73-6}


\bibitem[\protect\citeauthoryear{Seidman}{Seidman}{2013}]%
        {SeidmanPAN13:2013}
\bibfield{author}{\bibinfo{person}{Shachar Seidman}.}
  \bibinfo{year}{2013}\natexlab{}.
\newblock \showarticletitle{{Authorship Verification Using the Impostors Method
  Notebook for {PAN} at {CLEF} 2013}}. In \bibinfo{booktitle}{\emph{Working
  Notes for {CLEF} 2013 Conference , Valencia, Spain, September 23-26, 2013.}}
\newblock


\bibitem[\protect\citeauthoryear{Stamatatos}{Stamatatos}{2009}]%
        {StamatatosSurvey:2009}
\bibfield{author}{\bibinfo{person}{Efstathios Stamatatos}.}
  \bibinfo{year}{2009}\natexlab{}.
\newblock \showarticletitle{{A Survey of Modern Authorship Attribution
  Methods}}.
\newblock \bibinfo{journal}{\emph{J. Am. Soc. Inf. Sci. Technol.}}
  \bibinfo{volume}{60}, \bibinfo{number}{3} (\bibinfo{date}{March}
  \bibinfo{year}{2009}), \bibinfo{pages}{538--556}.
\newblock
\showISSN{1532-2882}


\bibitem[\protect\citeauthoryear{Stamatatos}{Stamatatos}{2013}]%
        {StamatatosRobustnessCharNGramsAA:2013}
\bibfield{author}{\bibinfo{person}{Efstathios Stamatatos}.}
  \bibinfo{year}{2013}\natexlab{}.
\newblock \showarticletitle{{On the Robustness of Authorship Attribution Based
  on Character N-Gram Features}}.
\newblock \bibinfo{journal}{\emph{Journal of Law and Policy}}
  \bibinfo{volume}{21} (\bibinfo{date}{01} \bibinfo{year}{2013}),
  \bibinfo{pages}{421--439}.
\newblock


\bibitem[\protect\citeauthoryear{Stamatatos}{Stamatatos}{2017}]%
        {StamatatosTextDistortion:2017}
\bibfield{author}{\bibinfo{person}{Efstathios Stamatatos}.}
  \bibinfo{year}{2017}\natexlab{}.
\newblock \showarticletitle{{Authorship Attribution Using Text Distortion}}. In
  \bibinfo{booktitle}{\emph{Proceedings of the 15th Conference of the European
  Chapter of the Association for the Computational Linguistics, {EACL} 2017,
  April 3-7, 2017, Valencia, Spain}}. \bibinfo{publisher}{The Association for
  Computer Linguistics}.
\newblock


\bibitem[\protect\citeauthoryear{Stamatatos, Daelemans, Verhoeven, Juola,
  L{\'{o}}pez{-}L{\'{o}}pez, Potthast, and Stein}{Stamatatos
  et~al\mbox{.}}{2015}]%
        {PANOverviewAV:2015}
\bibfield{author}{\bibinfo{person}{Efstathios Stamatatos},
  \bibinfo{person}{Walter Daelemans}, \bibinfo{person}{Ben Verhoeven},
  \bibinfo{person}{Patrick Juola}, \bibinfo{person}{Aurelio
  L{\'{o}}pez{-}L{\'{o}}pez}, \bibinfo{person}{Martin Potthast}, {and}
  \bibinfo{person}{Benno Stein}.} \bibinfo{year}{2015}\natexlab{}.
\newblock \showarticletitle{{Overview of the Author Identification Task at
  {PAN} 2015}}. In \bibinfo{booktitle}{\emph{Working Notes of {CLEF} 2015 --
  Conference and Labs of the Evaluation forum, Toulouse, France, September
  8--11, 2015}}. 17.
\newblock


\bibitem[\protect\citeauthoryear{Stamatatos, Daelemans, Verhoeven, Stein,
  Potthast, Juola, S{\'{a}}nchez{-}P{\'{e}}rez, and
  Barr{\'{o}}n{-}Cede{\~{n}}o}{Stamatatos et~al\mbox{.}}{2014}]%
        {PANOverviewAV:2014}
\bibfield{author}{\bibinfo{person}{Efstathios Stamatatos},
  \bibinfo{person}{Walter Daelemans}, \bibinfo{person}{Ben Verhoeven},
  \bibinfo{person}{Benno Stein}, \bibinfo{person}{Martin Potthast},
  \bibinfo{person}{Patrick Juola}, \bibinfo{person}{Miguel~A.
  S{\'{a}}nchez{-}P{\'{e}}rez}, {and} \bibinfo{person}{Alberto
  Barr{\'{o}}n{-}Cede{\~{n}}o}.} \bibinfo{year}{2014}\natexlab{}.
\newblock \showarticletitle{{Overview of the Author Identification Task at
  {PAN} 2014}}. In \bibinfo{booktitle}{\emph{Working Notes for {CLEF} 2014
  Conference, Sheffield, UK, September 15--18, 2014}}.
  \bibinfo{pages}{877--897}.
\newblock


\bibitem[\protect\citeauthoryear{Stamatatos, Fakotakis, and
  Kokkinakis}{Stamatatos et~al\mbox{.}}{2000}]%
        {StamatatosAutomaticTextCategorization:2000}
\bibfield{author}{\bibinfo{person}{Efstathios Stamatatos},
  \bibinfo{person}{Nikos Fakotakis}, {and} \bibinfo{person}{George~K.
  Kokkinakis}.} \bibinfo{year}{2000}\natexlab{}.
\newblock \showarticletitle{{Automatic Text Categorization in Terms of Genre
  and Author}}.
\newblock \bibinfo{journal}{\emph{Computational Linguistics}}
  \bibinfo{volume}{26}, \bibinfo{number}{4} (\bibinfo{year}{2000}),
  \bibinfo{pages}{471--495}.
\newblock


\bibitem[\protect\citeauthoryear{Stein, Lipka, and zu~Eissen}{Stein
  et~al\mbox{.}}{2008}]%
        {SteinMetaAnalysisAV:2008}
\bibfield{author}{\bibinfo{person}{Benno Stein}, \bibinfo{person}{Nedim Lipka},
  {and} \bibinfo{person}{Sven~Meyer zu Eissen}.}
  \bibinfo{year}{2008}\natexlab{}.
\newblock \showarticletitle{{Meta Analysis within Authorship Verification}}. In
  \bibinfo{booktitle}{\emph{19th International Workshop on Database and Expert
  Systems Applications {(DEXA} 2008), 1-5 September 2008, Turin, Italy}}.
  \bibinfo{publisher}{{IEEE} Computer Society}, \bibinfo{pages}{34--39}.
\newblock
\showISBNx{978-0-7695-3299-8}


\bibitem[\protect\citeauthoryear{Tax}{Tax}{2001}]%
        {TaxOCC:2001}
\bibfield{author}{\bibinfo{person}{David Martinus~Johannes Tax}.}
  \bibinfo{year}{2001}\natexlab{}.
\newblock \emph{\bibinfo{title}{{One-Class Classification: Concept Learning In
  the Absence of Counter-Examples}}}.
\newblock \bibinfo{thesistype}{Ph.D. Dissertation}. \bibinfo{school}{Delft
  University of Technology}.
\newblock
\showISBNx{9789075691054}


\bibitem[\protect\citeauthoryear{Veenman and Li}{Veenman and Li}{2013}]%
        {VeenmanPAN13:2013}
\bibfield{author}{\bibinfo{person}{Cor~J. Veenman} {and}
  \bibinfo{person}{Zhenshi Li}.} \bibinfo{year}{2013}\natexlab{}.
\newblock \showarticletitle{{Authorship Verification with Compression
  Features}}. In \bibinfo{booktitle}{\emph{Working Notes for {CLEF} 2013
  Conference , Valencia, Spain, September 23--26, 2013}}. 6.
\newblock


\end{thebibliography}
\end{document}